%% file: main.tex
\documentclass[letterpaper,11pt]{article}
\usepackage{etex}
\usepackage[margin=1in]{geometry}

\usepackage{amssymb,amsfonts,amsmath,amstext,amsthm,mathrsfs}
\usepackage{graphics,latexsym,epsfig,psfrag,wrapfig,comment,paralist}
\usepackage{xspace}
\usepackage{subcaption,bm,multirow}
\usepackage{booktabs}
\usepackage{mathdots}
\usepackage{bbold}
\usepackage{centernot}
\usepackage{prettyref}
\usepackage{listings}
\usepackage{tikz}
\usepackage{pgfplots}
\usetikzlibrary{shapes}
\usetikzlibrary{positioning, arrows, matrix, calc, patterns, fit}
\usepackage{caption}
\usepackage{array}
\usepackage{mdwmath}
\usepackage{multirow}
\usepackage{mdwtab}
\usepackage{eqparbox}
\usepackage{multicol}
\usepackage{amsfonts}
\usepackage{tikz}
\usepackage{multirow,bigstrut,threeparttable}
\usepackage{amsthm}
\usepackage{array}
\usepackage{bbm}
\usepackage{epstopdf}
\usepackage{mdwmath}
\usepackage{mdwtab}
\usepackage{eqparbox}
\usepackage{tikz}
\usepackage{latexsym}
\usepackage{amssymb}
\usepackage{bm}
\usepackage{graphicx}
\usepackage{mathrsfs}
\usepackage{epsfig}
\usepackage{psfrag}
\usepackage{setspace}
\usepackage[CJKbookmarks=true,
            bookmarksnumbered=true,
            bookmarksopen=true,
            colorlinks=true,
            citecolor=blue,
            linkcolor=blue,
            anchorcolor=red,
            urlcolor=blue
            ]{hyperref}
\usepackage{natbib}
\setcitestyle{authoryear,round}
\newtheoremstyle{break}
  {\topsep}{\topsep}%
  {\itshape}{}%
  {\bfseries}{}%
  {\newline}{}%

\newcommand{\calM}{{\mathcal{M}}}

\usepackage{pgfplots}
\usepackage[toc,title]{appendix}

\input{defs}

\newtheorem{remark}{\textbf{Remark}}[section]

\title{Provably Breaking the Quadratic Error Compounding Barrier in Imitation Learning, Optimally}
\author{Nived Rajaraman, Yanjun Han, Lin F. Yang, Kannan Ramchandran, Jiantao Jiao \thanks{Nived Rajaraman is with the Department of Electrical Engineering and Computer Sciences, University of California, Berkeley. Yanjun Han is with the Department of Electrical Engineering, Stanford University. Lin F. Yang is with the Electrical and Computer Engineering Department at the University of California, Los Angeles. Kannan Ramchandran is with the Department of Electrical Engineering and Computer Sciences, University of California, Berkeley. Jiantao Jiao is with the Department of Electrical Engineering and Computer Sciences and the Department of Statistics, University of California, Berkeley. Email: \{nived.rajaraman, jiantao, kannanr\}@eecs.berkeley.edu ; yjhan@stanford.edu; linyang@ee.ucla.edu.}}
\date{\today}

\begin{document}

\maketitle

\begin{abstract}
\noindent We study the statistical limits of Imitation Learning (IL) in episodic Markov Decision Processes (MDPs) with a state space $\mathcal{S}$. We focus on the known-transition setting where the learner is provided a dataset of $N$ length-$H$ trajectories from a deterministic expert policy and knows the MDP transition. We establish an upper bound $O(|\mathcal{S}|H^{3/2}/N)$ for the suboptimality using the \textsc{Mimic-MD} algorithm in~\citet{rajaraman2020fundamental} which we prove to be computationally efficient. In contrast, we show the minimax suboptimality grows as $\Omega( H^{3/2}/N)$ when $|\mathcal{S}|\geq 3$ while the unknown-transition setting suffers from a larger sharp rate $\Theta(|\mathcal{S}|H^2/N)$~\citep{rajaraman2020fundamental}. The lower bound is established by proving a two-way reduction between IL and the value estimation problem of the unknown expert policy under any given reward function, as well as building connections with linear functional estimation with subsampled observations. We further show that under the additional assumption that the expert is optimal for the true reward function, there exists an efficient algorithm, which we term as \textsc{Mimic-Mixture}, that provably achieves suboptimality $O(1/N)$ for arbitrary 3-state MDPs with rewards only at the terminal layer. In contrast, no algorithm can achieve suboptimality $O(\sqrt{H}/N)$ with high probability if the expert is not constrained to be optimal. Our work formally establishes the benefit of the expert optimal assumption in the known transition setting, while~\citet{rajaraman2020fundamental} showed it does not help when transitions are unknown. 
\end{abstract}

\tableofcontents

\input{introduction}

\input{main-contributions}
\input{known-transition-lowerbound}

\input{known-transition}

\clearpage

\bibliographystyle{plainnat}
\bibliography{refs}

\begin{appendices}
\input{appendix-A}

\end{appendices}

\end{document}

%% file: defs.tex
\usepackage{amsthm}

\usepackage{hyperref}
\hypersetup{
    colorlinks,
    linkcolor={blue},
    citecolor={red}
}
\usepackage{cleveref}

\usepackage{bbm}
\usepackage{algorithm}
\usepackage[noend]{algpseudocode}
\usepackage{caption}
\usepackage{subcaption}
\algnewcommand\algorithmicforeach{\textbf{for all}}
\algdef{S}[FOR]{ForAllDo}[1]{\algorithmicforeach\ #1\ }

\usepackage{xcolor}
\usepackage{tikz}
\usetikzlibrary{arrows,automata,positioning}

\DeclareMathOperator*{\argmin}{arg\,min}

\newtheorem{theorem}{Theorem}
\newtheorem{corollary}{Corollary}
\newtheorem{definition}{Definition}

\newtheorem{lemma}{Lemma}

\usepackage{hyperref}
\hypersetup{
    colorlinks,
    linkcolor={blue},
    citecolor={blue}
}
\usepackage{cleveref}

\usepackage{bbm}
\usepackage{algorithm}
\usepackage[noend]{algpseudocode}
\usepackage{caption}

\usepackage{xcolor}
\usepackage{tikz}
\usetikzlibrary{arrows,automata,positioning}

\usepackage{enumitem}

\usepackage{etoc}
\usepackage[toc,title]{appendix}
\usepackage{bm}

\usepackage[T1]{fontenc}    % use 8-bit T1 fonts
\usepackage{booktabs}       % professional-quality tables
\usepackage{amsfonts}       % blackboard math symbols
\usepackage{nicefrac}       % compact symbols for 1/2, etc.
\usepackage{microtype}      % microtypography
\usepackage{amsmath}
\usepackage{color}
\usepackage{setspace}

\newcommand{\calA}{{\mathcal{A}}}
\newcommand{\calS}{{\mathcal{S}}}

\def \bE {\mathbb{E}}
\newcommand{\Poi}{\mathsf{Poi}}
\newcommand{\calT}{{\mathcal{T}}}
\newcommand{\calG}{{\mathcal{G}}}

\usepackage{etex}

\usepackage{amsfonts,amsmath,amstext,mathrsfs}
\usepackage{graphics,latexsym,psfrag,wrapfig,comment,paralist}
\usepackage{xspace}
\usepackage{bm,multirow}
\usepackage{booktabs}
\usepackage{mathdots}
\usepackage{bbold}
\usepackage{centernot}
\usepackage{prettyref}
\usepackage{listings}
\usepackage{tikz}
\usepackage{pgfplots}
\usetikzlibrary{shapes}
\usetikzlibrary{positioning, arrows, matrix, calc, patterns, fit}
\usepackage{caption}
\usepackage{array}
\usepackage{mdwmath}
\usepackage{multirow}
\usepackage{mdwtab}
\usepackage{eqparbox}
\usepackage{multicol}
\usepackage{amsfonts}
\usepackage{tikz}
\usepackage{multirow,bigstrut,threeparttable}
\usepackage{array}
\usepackage{bbm}
\usepackage{epstopdf}
\usepackage{mdwmath}
\usepackage{mdwtab}
\usepackage{eqparbox}
\usepackage{tikz}
\usepackage{latexsym}
\usepackage{bm}
\usepackage{graphicx}
\usepackage{mathrsfs}
\usepackage{psfrag}
\usepackage{setspace}
\usepackage{natbib}
\setcitestyle{authoryear,round}

\usepackage{pgfplots}

\newtheorem*{insight*}{\textbf{Observation}}

\newtheorem*{proposition*}{\textbf{Proposition}}

\newtheorem*{lemmai*}{\textbf{Lemma (informal)}}

%% file: introduction.tex
\section{Introduction}

In many practical decision problems, it is difficult to design reward functions that accurately capture the task at hand. As mentioned in~\citet{Abbeel-Ng-ILviaIRL}, in autonomous driving it may be easy to state the reward function in English to be ``personal happiness'' or ``drive well'', but it is challenging to write down a specific reward function capturing this target mathematically. Often in practice, the reward function is manually refined \citep{ng1999policy,berner2019dota} until the learner demonstrates satisfactory behavior. This motivates the Imitation Learning (IL) problem which is the problem of learning from expert demonstrations in sequential decision making problems in the \textit{absence of reward feedback}. IL has found itself useful as a standalone framework~\cite{Abbeel-Ng-helicopter,Abbeel-Ng-ILviaIRL} as well as in combination with standard reinforcement learning workflows for policy optimization~\citep{salimans2018learning}. 

We study the imitation learning problem from a theoretical point of view in episodic Markov Decision Processes (MDPs) and introduce some notation before getting into the technical exposition. The \textit{value} $J_{\mathbf{r}} (\pi)$ of a (possibly stochastic) policy $\pi$ is defined as the expected cumulative reward accumulated over an episode of length $H$,
\begin{align} \label{eq:policy-value}
    J_{\mathbf{r}} (\pi) = \mathbb{E}_\pi \left[ \sum\nolimits_{t=1}^H \mathbf{r}_t (s_t,a_t) \right]
\end{align}
here $\mathbf{r}_t$ is the \emph{unknown} reward function of the MDP at time $t$, and the expectation $\mathbb{E}_\pi [\cdot]$ is computed with respect to the distribution over trajectories $\{ (s_1,a_1),\cdots,(s_H,a_H) \} $ induced by rolling out the policy $\pi= (\pi_1,\cdots,\pi_H)$ from some initial distribution $\rho(s)$, where $\pi_t$ denotes the policy at time step $t$. 
We assume that the reward functions $\mathbf{r}_t$ and MDP transitions $P_t$ could be time-variant and depend on $t$. We denote an MDP as a tuple $(\rho, P, H, \mathbf{r})$ throughout the paper, where $P=(P_1, P_2, \ldots, P_{H-1})$ and $\mathbf{r}=(\mathbf{r}_1, \mathbf{r}_2, \ldots \mathbf{r}_H)$. Notations $\Tilde{O}$ and $\Tilde{\Theta}$ omit logarithmic factors. We use $f_t^{\pi}$ to denote the state distribution induced at time $t$ by the policy $\pi$. In the most basic setting of IL, the learner is provided a dataset of $N$ trajectories rolling out an unknown expert policy $\pi^*$. The objective of the learner is to construct a policy $\widehat{\pi}$ with provably \emph{small} \emph{suboptimality}, which is a random variable defined as the difference in value of the expert's and learner's policies: $J_{\mathbf{r}} (\pi^*) - J_{\mathbf{r}} ( \widehat{\pi})$, where $\mathbf{r}$ represents the ``true'' reward function we care about. In this paper we assume the expert policy $\pi^*$ is deterministic\footnote{Stochastic policies exhibit fundamentally different behaviors as shown in~\citep{rajaraman2020fundamental}.}.

One of the most natural approaches towards imitation learning is Behavior Cloning (BC)~\citep{Ross-AIstats10}, which reduces the IL problem to a supervised learning problem with state being feature and action being labels. It was shown in~\citep{Ross-AIstats10} that if the \emph{population} supervised learning 0-1 loss is $\epsilon$, then the suboptimality in IL is $O(\epsilon H^2)$, where the factor $H^2$ increase is called \emph{error compounding}. \citet{Ross-AIstats10} showed that this $H^2$ reduction relationship is not improvable for BC\footnote{Precisely, there exists some MDP such that the worst case ratio of the expected suboptimality in IL and supervised learning 0-1 loss is at least $\Omega(H^2)$.}, but it remained unclear whether there exists an information theoretic lower bound showing that the $H^2$ factor is inevitable for \emph{all} algorithms. Indeed, the sample size $N$ does not appear in the reduction step in BC~\citep{Ross-AIstats10}, and in the lower bound instance of~\citep{Ross-AIstats10} the learner only needs a single trajectory of observations to achieve suboptimality zero. Recently, \citet{rajaraman2020fundamental} studied this problem in a statistical setting and showed that \emph{any} algorithm has to suffer from suboptimality $|\mathcal{S}|H^2/N$ in the IL setting in the worst case, and BC achieves this bound for deterministic expert demonstrations with the supervised learning loss being $O(|\mathcal{S}|/N)$. Moreover, the $|\mathcal{S}|H^2/N$ lower bound still holds even if the learner can interactively query the expert as in the setting of DAGGER~\citep{RossGB11} and when the expert is assumed to be \emph{optimal} for the true reward function.

Hence, it seems natural to impose either structural assumptions on the MDP or change the sampling model to mitigate the $H^2$ error compounding effect. \citet{rajaraman2020fundamental} took the second approach and assumed that the initial state distribution and the Markov transition kernels are completely known to the learner. To motivate this setting, we first provide an intuitive understanding of the lower bound instance in~\citet{rajaraman2020fundamental} which shows no algorithm can beat the $|\mathcal{S}|H^2/N$ lower bound in the no-interaction setting\footnote{Similar intuitive explanations have appeared in the literature, and there has been no information theoretic lower bound proof before~\cite{rajaraman2020fundamental}.}. In a nutshell, without transition information the learner cannot recover after making a ``mistake'' via playing an action different from the expert's action at some state. Concretely, in the lower bound instance~\citep{rajaraman2020fundamental}, at each time $t$ any learner has a $\frac{|\mathcal{S}|}{N}$ probability of making a mistake due to not having observed the expert action at this time step\footnote{The term $|\mathcal{S}|/N$ is an upper bound on the \emph{missing mass} in sampling~\cite{McAllesterOrtiz}, which can also be understood as the $V/N$ regret in binary classification with zero oracle error, where $V = |\mathcal{S}|$ is the VC-dimension when $|\mathcal{A}| = 2$. }. Then, they make the probability that the learner has made a mistake up to time $t$ to be $\asymp \frac{|\mathcal{S}| t}{N}$ by setting the union bound to be tight, and under this event the learner incurs a suboptimality of $1$. Thus, the expected suboptimality of the learner is $ \sum_{t=1}^H \frac{|\mathcal{S}| t }{N} \asymp \frac{|\mathcal{S}|H^2}{N}$.

To break the the quadratic dependence, the analysis above suggests that one needs to beat the union bound $|\mathcal{S}|t/N$ at time $t$, which implies that one needs to conduct \emph{long-range planning}: the error events of making mistakes at each time $t$ should be made \emph{highly negatively correlated} in the sense that we can quickly recover from the mistakes we made in the past. Intuitively, in autonomous driving, recovering from a mistake of going off the road corresponds to trying to go back to the main road, and this step requires the knowledge of the MDP transitions. 

Clearly, in a simulation environment, or model-based reinforcement learning setting it is reasonable to assume that we have a good knowledge about the model, but more interestingly many practical algorithms~\citep{ho2016generative,fu2017learning,Brantley2020Disagreement-Regularized} can be viewed as gradient and/or sample based methods to approximately solve optimization problems defined in the known transition setting.

In this setting, \citet{rajaraman2020fundamental} proposed an algorithm called \textsc{Mimic-MD} and show that its suboptimality is upper bounded by $|\mathcal{S}| H^{3/2}/N$, which provably breaks the $H^2$ dependence. However, there are several key problems that remain unsolved:
\begin{enumerate}
    \item[(i)] \textsc{Mimic-MD} has suboptimality upper bounded by $|\mathcal{S}| H^{3/2}/N$, but the best known lower bound in the known transition setting~\citep{rajaraman2020fundamental} is $|\mathcal{S}|H/N$, which does not match. What is the dependence of the minimax rate on the horizon $H$?
    \vspace{-1.5mm}
    \item[(ii)] Is \textsc{Mimic-MD} in~\citep{rajaraman2020fundamental} efficiently computable?
    \vspace{-1.5mm}
    \item[(iii)] Furthermore, we know that in the basic setting assuming that the expert is an optimal policy on the underlying true reward function does not help in the worst case: the minimax suboptimality is still $\asymp |\mathcal{S}|H^2/N$~\citep{rajaraman2020fundamental}, but does knowing the expert is optimal help in the known transition setting? 
    \vspace{-1.5mm}
    \item [(iv)] How can we connect algorithms developed in the known transition setting such as \textsc{Mimic-MD} to practical imitation learning algorithms? What practical insights do these theoretical analysis provide?  
    \vspace{-1.5mm}
\end{enumerate}

%% file: main-contributions.tex
\subsection*{Main Contributions}

In this paper, we make the following contributions. 
\begin{enumerate}
\vspace{-1.5mm}
    \item [(i)] We show that the optimal dependence on the horizon for imitation learning in the known transition setting when all bounded rewards are considered is indeed $H^{3/2}$, and show that any algorithm has to suffer from suboptimality $H^{3/2}/N$ when $|\mathcal{S}| \ge 3$ (Theorem~\ref{theorem:lb-known-transition}). We also show that the \textsc{Mimic-MD} algorithm in~\citep{rajaraman2020fundamental} can be efficiently solved in polynomial time, whose statistical performance achieves the bound $|\mathcal{S}|H^{3/2}/N$~(Theorem~\ref{thm.mimimdefficient}). Moreover, we demonstrate a gap between binary state space $|\mathcal{S}| = 2$ and larger state space $|\mathcal{S}|\geq 3$: when $|\mathcal{S}| = 2$, the minimax rates become $\widetilde{\Theta}(H/N)$, and \textsc{Mimic-MD} also achieves the minimax rate~(Theorem~\ref{thm.mimimdefficient}). 
    \vspace{-1.5mm}
    \item [(ii)] In our lower bound proof for the minimax rates above, we constructed a special family of MDPs, which we term \emph{4-state MDPs}~(Section~\ref{sec.4statemdp}), that precisely exploit the vulnerabilities of any imitation learning algorithms. To make the results stronger, the 4-state MDP in fact has \emph{time-invariant} Markov transition functions and rewards, but it still appears to be the most difficult case even among the family of MDPs with possibly
    \emph{time-variant} transitions and rewards. In particular, we show that in the 4-state MDP instance, IL is no easier than estimating a linear functional whose parameters are subsampled with constant fractions, whose statistical fundamental limit proof may be of independent interest. 
    \vspace{-1.5mm}
    \item [(iii)] We show that if one imposes the additional assumption that the expert policy is \emph{optimal} for the true reward function we use the evaluate the suboptimality gap, then we can construct an \emph{efficient} algorithm that provably achieves suboptimality $\widetilde{O}(H/N)$ for the 4-state MDP for \emph{any} expert policy~(Theorem~\ref{thm.4statemdpunbiased}). The result can be easily extended to nearly optimal policies. 
    \vspace{-1.5mm}
    \item [(iv)] We generalize the efficient algorithm that achieves linear in $H$ suboptimality for the 4-state MDP to \emph{arbitrary} 3-state MDPs, and show that in the case where only the terminal state has rewards, which is a common case in a variety of reinforcement learning applications, one can construct an efficient algorithm that achieves suboptimality $\widetilde{O}(1/N)$, without dependence on $H$ at all~(Corollary~\ref{cor.3stategeneral}). We call the algorithm \textsc{Mimic-Mixture}~(Algorithm~\ref{alg.minimcmixture}). This is precisely the reason why in the 4-state MDP the dependence on $H$ is linear, since every time step only contributes $1/N$. 
    \vspace{-1.5mm}
    \item [(v)] We propose a framework of reductions between imitation learning and (uniform) expert value estimation~(Section~\ref{sec.equivalence}), and propose a general minimax optimization framework in known transition setting, which subsumes various practical algorithms in practice~\citep{ratliff2006maximum,ho2016generative,fu2017learning}, and show that \textsc{Mimic-MD} can be viewed as special cases of this framework. Combined with our theoretical results, GAIL~\citep{ho2016generative} corresponds to~\eqref{eq:value-learner} using the empirical estimator~(\ref{eqn.simpletv}), while AIRL~\citep{fu2017learning} corresponds to utilizing the expert optimal information. Our work implies that algorithm exploiting the expert optimal assumption could have significantly superior performances in practice for long horizons. 
    \vspace{-1.5mm}
    \item [(vi)] We draw connections between IL and online reinforcement learning. In IL, we have expert demonstrations, no reward information; in online RL, we have reward information, but need to decide the actions.  Correspondingly, the IL minimax rate is $|\mathcal{S}|H^2/N$~\citep{rajaraman2020fundamental} while the RL minimax rate is $\sqrt{H^3 |\mathcal{S}||\mathcal{A}|/N}$~\citep{domingues2020episodic}\footnote{Here we do not compare IL and RL in the known transition setting since known transition is rarely considered in RL.}. Interpreting the quantities in terms of sample complexities, the RL results are uniformly worse in terms of dependence on $|\mathcal{A}|, H, \epsilon$, which conveys an interesting conceptual message: it is much better to have a good teacher instead of myopic local reward information. We also emphasize that without operating in a statistical framework, it would have been impossible to compare IL and online RL, and a consistent theoretical framework provides a unified benchmark for comparing different formulations. 
    \vspace{-1.5mm}
\end{enumerate}

\section{Related work}

The imitation learning problem has been extensively studied in the literature~\cite{Abbeel-Ng-ILviaIRL, syed2008apprenticeship, ratliff2006maximum, ziebart2008maximum,Ross-AIstats10,finn2016guided, fu2017learning, pan2017agile,ke2019imitation}, and various algorithms~\citep{ho2016generative,laskey2017dart,Luo2020Learning,Brantley2020Disagreement-Regularized,zhang2020generative} have been proposed to overcome the $H^2$ error compounding issue defined by~\citep{Ross-AIstats10}, and some require actively querying the expert during training~\citep{RossGB11,Ross2014ReinforcementAI,sun2017deeply}. The majority of literature do not focus on obtaining minimax rates in statistical settings, and we emphasize that regret results obtained in the online learning framework~\citep{RossGB11,lee2018dynamic} may not imply equally strong statistical results since it is difficult to evaluate the performance of the oracle term in the regret formulation. Indeed, it was shown in~\citep{rajaraman2020fundamental} that even in the interactive IL setting where the expert can be actively queried, \emph{no} algorithm can beat the $|\mathcal{S}|H^2/N$ information theoretic lower bound. The literature has also studied the combination of representation learning and imitation learning~\cite{arora2020provable} as well as IL without expert action information~\citep{nair2017combining,  torabi2018behavioral,wen-sun-ILFO,arora2020provable}. Recently, \citet{xu2020error} studied statistical performances of BC and GAIL~\citep{ho2016generative} and showed that GAIL could improve the horizon dependence from $1/(1-\gamma)^2$ to $1/(1-\gamma)$. Again, the lower bound argument in~\citep{xu2020error} was based on reduction and not information theoretic, and its analysis of GAIL implicitly assumes known transition\footnote{In practice, we would query the environment to obtain the MDP transition functions to approximately simulate the known transition setting. } whose rate degrades to $1/\sqrt{N}$ compared to $1/N$ in BC. We believe one can export our analysis to the discounted case to obtain similar results.

%% file: known-transition-lowerbound.tex
\section{Minimax algorithmic framework and reductions between IL and expert value estimation} \label{sec.equivalence}

We first define the IL problem under high probability:
\begin{definition}[Imitation learning]
Given a collection of MDP instances $\{(\rho, P, H, \mathbf{r}, \pi^*)\}$, we say that an algorithm output $\widehat{\pi}$ solves the imitation learning problem with confidence $1-\delta$ and error $\epsilon$ if for any such instance $(\rho, P, \mathbf{r}, \pi^*)$, we have
\begin{align*}
    \mathbb{P}_{\mathcal{D}(\rho, P, \pi^*)}(J_{\mathbf{r}}(\pi^*) -J_{\mathbf{r}}(\widehat{\pi}) \geq \epsilon)\leq \delta,
\end{align*}
where the algorithm observes $\rho$ and $P$, and a dataset $\mathcal{D}(\rho, P, \pi^*)$ of the expert trajectories (no rewards), but not the expert policy $\pi^*$ or $\mathbf{r}$ directly.
\end{definition}

We then define the problem of expert value estimation:
\begin{definition}[Expert value estimation]
Given a collection of MDP instances $\{(\rho, P, H, \mathbf{r}, \pi^*)\}$, we say that an estimator $\widetilde{J}_{\mathbf{r}} (\pi^*)$ is an estimator for $J_{\mathbf{r}}(\pi^*)$ with confidence $1-\delta$ and error $\epsilon$ if for any such instance $(\rho, P, \mathbf{r}, \pi^*)$, we have
\begin{align*}
    \mathbb{P}_{\mathcal{D}(\rho, P, \pi^*)}(|J_{\mathbf{r}}(\pi^*) -\widetilde{J}_{\mathbf{r}} (\pi^*)| \geq \epsilon)\leq \delta.
\end{align*}
where the estimator $\widetilde{J}$ is a function of $\rho$, $P$, $\mathbf{r}$ and a dataset $\mathcal{D}(\rho, P, \pi^*)$ of the expert trajectories (no rewards), but not the expert $\pi^*$ directly.
\end{definition}

Correspondingly, the problem of \emph{uniform} expert value estimation is defined as:
\begin{definition}[Uniform expert value estimation]\label{def.uniformvalueestimation}
Given a collection of  instances $\{(\rho, P, H, \mathbf{r}, \pi^*)\}$, we say that an estimator $\widetilde{J}_{\mathbf{r}} (\pi^*)$ is an estimator for $J_{\mathbf{r}}(\pi^*)$ with confidence $1-\delta$ and error $\epsilon$ if for any such instance $(\rho, P, \mathbf{r}, \pi^*)$, we have
\begin{align*}
    \mathbb{P}_{\mathcal{D}(\rho, P, \pi^*)}(\sup_{\mathbf{r}\in \mathcal{R}_D}|J_{\mathbf{r}}(\pi^*) -\widetilde{J}_{\mathbf{r}} (\pi^*)| \geq \epsilon)\leq \delta,
\end{align*}
where the estimator $\widetilde{J}$ is function of $\rho$, $P$, $\mathbf{r}$, a dataset $\mathcal{D}(\rho, P, \pi^*)$ of the expert trajectories (no rewards), and an input set of reward functions, $\mathcal{R}_D$, which also contains the true reward, but not the expert $\pi^*$ directly. We added the subscript $D$ to emphasize that this set $\mathcal{R}_D$ could depend on the data seen by the learner, but we may omit it when clear from the context. 
\end{definition}

We propose a general minimax formulation to reduce the IL problem to uniform expert value estimation with known transitions. Define the learner policy $\widehat{\pi}$ as the solution to the following minimax optimization problem:
\begin{equation} \label{eq:value-learner}
    \widehat{\pi} \gets \argmin_{\pi} \max_{\mathbf{r} \in \mathcal{R}_D} \widetilde{J}_{\mathbf{r}} (\pi^*) - J_{\mathbf{r}} (\pi)\tag{\textsf{OPT}},
\end{equation}
where $\mathcal{R}_D$ is the same as that in Definition~\ref{def.uniformvalueestimation}. The next result shows reductions between IL and expert value estimation when transitions are known. 
\begin{theorem}[Reductions between IL and expert value estimation with known transitions]\label{thm.reductions}
Consider the following two cases of $\mathcal{R}_D$:
\begin{enumerate}
    \vspace{-1mm}
    \item[(i)] reward function symmetric: for any $(\rho, P, \mathbf{r}, \pi^*)$ instance we consider,  $(1-\mathbf{r})$ is also in the set. A notable special case is when we consider all possible reward functions bounded between zero and one;
    \vspace{-1.5mm}
    \item[(ii)] expert optimal: for each $(\rho, P, \mathbf{r}, \pi^*)$ instance, $\pi^*$ is an optimal policy for $\mathcal{M}=(\rho, P, H, \mathbf{r})$.  
    \vspace{-1.5mm}
\end{enumerate}
Then, under both cases, 
\begin{enumerate}
    \vspace{-1mm}
     \item[(i)] if $\widehat{\pi}$ solves IL with confidence $1-\delta$ and error $\epsilon$, then $J_{\mathbf{r}}(\widehat{\pi})$ solves expert value estimation with confidence $1-2\delta$ and error $\epsilon$;
     \vspace{-1.5mm}
     \item[(ii)] if $\widetilde{J}_{\mathbf{r}} (\pi^*)$ solves uniform expert value estimation with confidence $1-\delta$ and error $\epsilon$, then the minimax algorithm in \eqref{eq:value-learner} solves IL with confidence $1-\delta$ and error $2\epsilon$. 
     \vspace{-1.5mm}
\end{enumerate}
\end{theorem}

\section{Minimax Bounds for General Expert}

For brevity we use $\mathcal{R}$ to denote the set of all reward functions such that $\mathbf{r}_t(s,a)\in [0,1]$. Applying the conclusion in Section~\ref{sec.equivalence}, we can solve IL via constructing a uniform expert value estimator such that $\sup_{\mathbf{r}\in \mathcal{R}} |J_{\mathbf{r}}(\pi^*) - \widetilde{J}_{\mathbf{r}}(\pi^*)| $ is small. If we denote the marginal distribution of $(S_t,A_t)$ when we roll out the policy $\pi^*$ as $f_t^{\pi^*}(s,a)$, we can rewrite $J_{\mathbf{r}}(\pi^*) = \sum_{t = 1}^H \mathbb{E}_{(S,A) \sim f_t^{\pi^*}}[r_t(S,A)]$. Hence it motivates us to estimate $J_{\mathbf{r}}(\pi^*)$ using $\sum_{t = 1}^H \mathbb{E}_{(S,A) \sim \widehat{f}_t^{\pi^*}}[r_t(S,A)]$, where $\widehat{f}_t^{\pi^*}$ is some estimator of $f_t^{\pi^*}$. In this case, the uniform value estimation error reduces to the sum of \emph{total variation} distances $ \sum_{t=1}^H \mathsf{TV}(f_t^{\pi^*},\widehat{f}_t^{\pi^*})$, where we used the assumption that $\mathsf{TV}(P,Q) = \sup_{r\in [0,1]^{\mathcal{S}\times\mathcal{A}}} \mathbb{E}_P[r] - \mathbb{E}_Q[r]$. 

For each $t$, we obtain $N$ i.i.d. samples from distribution $f_t^{\pi^*}$, and it is natural to use the empirical distribution as the estimator $\widehat{f}_t^{\pi^*}$. It then follows from standard results~\citep{han2015minimax} that with probability at least $1-\delta$, we have \footnote{Note that we have assumed the expert policy is deterministic, so the support of $f_t^{\pi^*}$ is at most $|\mathcal{S}|$. }
\begin{align}\label{eqn.simpletv}
    \sum_{t=1}^H \mathsf{TV}(f_t^{\pi^*},\widehat{f}_t^{\pi^*}) \lesssim H \sqrt{\frac{|\mathcal{S}| + \log(H/\delta)}{N}}. 
\end{align}
Although~(\ref{eqn.simpletv}) seems to suggest that the suboptimality error now grows linear in $H$, there is a catch: the dependence on $N$ has degraded to $N^{-1/2}$, and this bound becomes even worse than the behavior cloning result $|\mathcal{S}|H^2/N$ when $N$ is large. 

The advance made in~\citet{rajaraman2020fundamental} about the \textsc{Mimic-MD} algorithm is that there exists an \emph{improved} estimator for $f_t^{\pi^*}$ that achieves smaller $\mathsf{TV}$ loss. Indeed, the empirical distribution $\widehat{f}_t^{\pi^*}$ does not even utilize the transition information! Intuitively, given the transition information, one may simulate infinitely many new trajectories and view them as new data to improve the statistical efficiency. The only case where simulation fails would be that we encounter a state where we have not visited in the observations, but the probability of seeing an unseen state within the first $t$ steps is at most $|\mathcal{S}|t/N$ by union bound, hence the total variation loss in estimating $f_t^{\pi^*}$ can be improved to be $\sqrt{|\mathcal{S}|/N} \sqrt{|\mathcal{S}|t/N} = \frac{|\mathcal{S}|\sqrt{t}}{N}$.\footnote{Precisely, the $L_1$ error of estimating a discrete distribution $(p_1,p_2,\ldots,p_k)$ from $N$ i.i.d. samples is upper bounded by $\sum_{i\in [k]}\sqrt{\frac{p_i}{N}} \leq \sqrt{k/N}$ where the worst case is attained when $p_i \equiv 1/k$. In \textsc{Mimic-MD}, we are reducing the effective probability mass from $1$ to $\frac{|\mathcal{S}|t}{N}$, so the problem is reduced to upper bounding $\sup_{p_i\geq 0, \sum_{i\in [k]}p_i\leq \frac{|\mathcal{S}|t}{N}} \sum_{i} \sqrt{\frac{p_i}{N}} = \sqrt{\frac{k}{N}} \sqrt{\frac{|\mathcal{S}|t}{N}}$. }
Summing up $\sum_{t = 1}^H \frac{|\mathcal{S}|\sqrt{t}}{N} \lesssim |\mathcal{S}|H^{3/2}/N$. 
The following theorem summarizes the performance of \textsc{Mimic-MD}, which is the specific  instantiation of ~\eqref{eq:value-learner} when we consider all possible rewards and the the improved value estimator mentioned above. 

\begin{theorem}\label{thm.mimimdefficient}
The optimization problem \textsc{Mimic-MD} in \eqref{eq:opt} can be formulated as a convex program and is efficiently solvable in $\text{poly}(|\mathcal{S}|,|\mathcal{A}|,H)$ time. Moreover, its solution $\widehat{\pi}$ achieves expected suboptimality gap
\begin{align*}
    J_{\mathbf{r}}(\pi^*) - \mathbb{E}[J_{\mathbf{r}}(\widehat{\pi})] \lesssim \begin{cases} H \log H/N & |\mathcal{S}| = 2 \\ |\mathcal{S}| H^{3/2}/N & |\mathcal{S}|\geq 3  \end{cases}
\end{align*}
for all MDPs with state space $\mathcal{S}$, action space $\mathcal{A}$, horizon $H$, and rewards bounded $\mathbf{r}_t \in [0,1]$. 
\end{theorem}

Theorem~\ref{thm.mimimdefficient} points out that when $|\mathcal{S}| = 2$, the expected suboptimality achieved by \textsc{Mimic-MD} is in fact nearly linear in $H$, which nearly matches the lower bound in \citep{rajaraman2020fundamental}\footnote{Indeed, whenever we have an unseen state for a distribution with binary values, the total variation distance between the empirical distribution and the real distribution is of order $\widetilde{O}(1/N)$. However, it is not true for distributions with $|\mathcal{S}|\geq 3$ in general: for example, if $p = (1/2,1/2-1/N,1/N)$, then with constant probability we will not see the third state in the dataset, but the total variation distance between empirical distribution and true distribution scales as $O(1/\sqrt{N})$. }.

What is the optimal dependence on $H$ when $|\mathcal{S}|\geq 3$? We show in the following result, that indeed this $H^{3/2}$ dependence is a barrier for any learner.

\begin{theorem} \label{theorem:lb-known-transition}
Suppose $H \ge 2$ and $N \ge 7$. If $N \ge 6H$, for every learner $\widehat{\pi}$, there exists an MDP $\mathcal{M}$ on $3$ states such that,
\begin{equation*}
    \mathrm{Pr} \left( J_{\mathcal{M}} (\pi^*) - J_{\mathcal{M}} (\widehat{\pi}) \ge \frac{c H^{3/2}}{N} \right) \ge c',
\end{equation*}
for some constants $c, c' > 0$. Here the probability is computed over the randomness of the dataset $D$ as well as the external randomness employed by $\widehat{\pi}$.
\end{theorem}

\subsection{The 4-state MDP example}\label{sec.4statemdp}

To best illustrate the insights behind the lower bound construction, we construct a particular lower bound instance, which we name \emph{4-state MDP}, and describe informally the main ideas. In Appendix~\ref{sec.lowerbound-knowntr}, we formally prove the lower bound for $|\mathcal{S}| = 3$ as well by essentially combining the state labelled $1$ with $3$ and $4$.

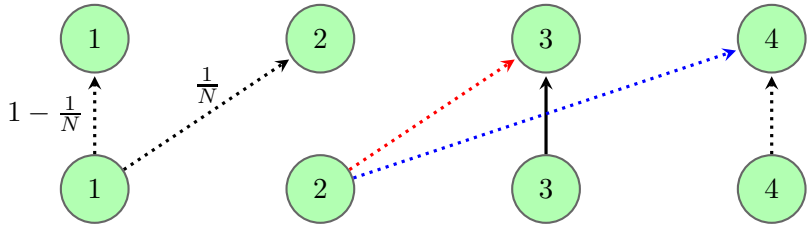
\begin{figure}[htb!]
\centering
\begin{tikzpicture}[shorten >=1pt,node distance=1.0cm,on grid,auto,good/.style={circle, draw=black!60, fill=green!30!white, thick, minimum size=9mm, inner sep=0pt},bad/.style={circle, draw=black!60, fill=red!30, thick, minimum size=9mm, inner sep=0pt}]
\tikzset{every edge/.style={very thick, draw=black}}
    
\node[good]     (s1)                   {$1$};
\node[good]     (s2) [right = 3cm of s1]  {$2$};
\node[good]     (s3) [right = 3cm of s2]  {$3$};
\node[good]     (s4) [right = 3cm of s3]  {$4$};

\node[good]     (s1new) [above = 2cm of s1]     {$1$};
\node[good]     (s2new) [right = 3cm of s1new]  {$2$};
\node[good]     (s3new) [right = 3cm of s2new]  {$3$};
\node[good]     (s4new) [right = 3cm of s3new]  {$4$};

\path[->,>=stealth]
(s1)    edge [dotted] node [left] {$1-\frac{1}{N}$} (s1new)
(s1)    edge [dotted] node [above] {$\frac{1}{N}$} (s2new)
(s2)    edge [red, dotted] node {} (s3new)
(s3)    edge [] node {} (s3new)
(s4)    edge [dotted] node {} (s4new)
(s2)    edge [blue, dotted] node {} (s4new);

\end{tikzpicture}
\caption{The \emph{4-state MDP}: Dotted (resp. solid) lines indicate transitions that provide a reward of $0$ (resp. $1$). The states $1$, $3$ and $4$ have a single action, with the transition probabilities indicated above the arrow. On the other hand, state $2$ has $2$ actions: with probability $1$, the red action transitions the learner to state $3$ while the blue action transitions the learner to state $4$.}
\label{fig:4-state-MDP}
\end{figure}

Since Theorem~\ref{thm.reductions} shows that value estimation is not harder than IL, it suffices to show that the value estimation error is at least $H^{3/2}/N$. Concretely, the $4$-state MDP in \Cref{fig:4-state-MDP} is time-invariant with the states labelled $1$, $2$, $3$ and $4$. All states besides $2$ are trivial and without loss of generality have only a single action. The state $2$ has exactly $2$ actions, one leading deterministically to state $3$ and the other to state $4$. Furthermore, the reward function of the MDP is all $1$ on the state $3$ (recall there is only a single action at this state). We assume the initial distribution is $(1-1/N,1/N,0,0)$. 

Define variables
\begin{align*}
    U_i & = \begin{cases} 1 & \text{ if } \pi^*_i(\text{red}\mid 2) = 1 \\ 0  & \text{ if } \pi^*_i(\text{blue}\mid 2) = 1\end{cases}
\end{align*}
Note that the marginal distribution at state $1$ at time $t$ is independent of expert policy and equal to $(1-1/N)^t$, and the marginal distribution of state $2$ at time $t$ is always $w_t \triangleq (1-1/N)^{t-1}/N$. Consider the case that $N\gtrsim H$, in this case the marginal probability of state $2$ for every time step is $w_t \asymp 1/N$. We say that the contribution of time $t$ to final expert value is $v_t = \sum_{i = 1}^{t-1} w_i U_i$, and the final \emph{expert value} $V^*$ is given by
\begin{align}\label{eqn.vstarexpression}
    V^* & = \sum_{t = 1}^{H} v_t,
\end{align}
and we aim to show the estimation error of $V^*$ is at least $\asymp H^{3/2}/N$. 

The major step towards the final lower bound is to show that the estimation error of $v_H$, which is the contribution to the final value from the last layer, is at least $\sqrt{H}/N$. This dependence on the time horizon would then accumulate to achieve the $H^{3/2}$ result. Note that 
$
    v_H = \sum_{t = 1}^{H-1} w_t U_t,
$
can be viewed as the weighted combination of parameters $U_i$, but since the marginal probability of state $2$ is about $1/N$, in total $N$ trajectories there will be a constant fraction of state $2$'s across time steps that are not observed in the dataset. If we impose a uniform prior on $U_i \sim \mathsf{Bern}(1/2)$, then the posterior variance of $v_H$ is at least a constant fraction of the prior variance of $v_H$, which is 
\begin{align*}
    \mathsf{Var}(v_H) & = \sum_{t= 1}^{H-1} \frac{w_t^2}{4} \asymp \frac{H}{N^2},
\end{align*}
which implies that the posterior standard deviation of $v_H$ is at least of order $\sqrt{H}/N$. Then, we can combine this lower bound with~(\ref{eqn.vstarexpression}) to show that the overall estimation error of $V^*$ is at least $\sum_{t= 1}^H \sqrt{t}/N \asymp H^{3/2}/N$. 

This result implies that \textsc{Mimic-MD} indeed achieves optimal dependence on the MDP horizon $H$, growing as $H^{3/2}$.

%% file: known-transition.tex
\section{Known transition and expert optimal} \label{section:known-transition}

Does the lower bound in Theorem~\ref{theorem:lb-known-transition} still hold when we impose the additional assumption that the expert policy $\pi^*$ is the optimal policy \footnote{Our theory can be easily extended to near-optimal policies.} for the true reward function $\mathbf{r}$? Attempting to deploy the $4$-state MDP instance with the expert optimal assumption, we encounter the following difficulties:
\begin{enumerate}
\vspace{-1mm}
    \item [(i)] if we follow the current proof and only put rewards on state $3$, then the optimal policy would be only choosing the \emph{red} action, hence the posterior uncertainty of $v_H$ would not scale with $H$ since with a single action at state $2$ would reveal the whole policy $\pi^*$;
    \vspace{-1.5mm}
    \item [(ii)] if we put rewards on the links from 2 to 3 if $U_i = 1$ and from 2 to 4 if $U_i = 0$, then we can still impose the uniform Bernoulli priors on $\{U_i\}_{i=1}^H$, but the posterior standard deviation of $v_H$ would still be $1/N$ since in this case only state 2 would contribute to the value but its marginal probability is independent of $\pi^*$ and always $\asymp 1/N$. 
    \vspace{-1mm}
\end{enumerate}

It begs the question: can we formally prove that for the 4-state MDP instance, for \emph{any} policy $\pi^*$, if we assume it is optimal for the true reward function $\mathbf{r}$, can we show that the suboptimality $J_{\mathbf{r}}(\pi^*) - J_{\mathbf{r}}(\widehat{\pi}) \lesssim \frac{H \log (NH)}{N}$ with probability 0.99? 

The crucial observation we make here, is that it suffices to find a policy $\widehat{\pi}$ such that its \emph{expected} suboptimality is small to guarantee suboptimality small with constant probability. Indeed, if we can show
$
    J_{\mathbf{r}}(\pi^*) - \mathbb{E}[J_{\mathbf{r}}(\widehat{\pi})] = \mathbb{E}[J_{\mathbf{r}}(\pi^*) -J_{\mathbf{r}}(\widehat{\pi})]
$
is small, it immediately implies a concentration bound on $J_{\mathbf{r}}(\pi^*) -J_{\mathbf{r}}(\widehat{\pi})$ using Markov's inequality thanks to our assumption that $\pi^*$ is optimal for the reward function $\mathbf{r}$. 

To achieve small expected suboptimality, since $\mathbf{r}$ is deterministic, it suffices to find some policy $\widehat{\pi}$ whose expected state-action occupancy measure is close to that of the expert policy. We remark that the unbiased \emph{estimation} of the probability $\mathrm{Pr}_{\pi^*}(s_H = s^*)$ is in fact trivial and achieved by the empirical distribution of the state $s^*$; however, our target of \emph{realization} is much more difficult and requires to achieve a small bias using some policy $\widehat{\pi}$. For example, one of the key challenges in proving Theorem~\ref{thm.4statemdpunbiased} and~\ref{theorem:single_state} is that the empirical distribution of the state $s^*$ may not be achievable by any policy owing to the possibly limited approximation power of the MDP. The next theorem shows we can solve the 4-state MDP instance with nearly linear dependence on $H$. 

\begin{theorem}\label[theorem]{thm.4statemdpunbiased}
Given observed $N$ trajectory rollouts, there exists an efficient algorithm to compute a policy $\widehat{\pi}$ such that for the 4-state MDP instance,
\begin{align}
    J_{\mathbf{r}}(\pi^*) -J_{\mathbf{r}}(\widehat{\pi}) \lesssim \frac{H \log (NH)}{N}
\end{align}
with probability 0.99 for any $\mathbf{r}$ such that $\pi^*$ is optimal. 
\end{theorem}

We defer the proof of Theorem~\ref{thm.4statemdpunbiased} to Appendix~\ref{app.4stateunbiasedproof}, but present the explicit policy construction for the single state 3 at layer $H$ here, which conveys the key insights of the algorithm. Let $X_t$ be the number of trajectories in which the expert visits state $2$ at time $t$ in the dataset and the $X_t$'s are jointly following a multinomial distribution. The policy $\widehat{\pi} = (\widehat{\pi}_1, \widehat{\pi}_2,\ldots,\widehat{\pi}_H)$ we output is:
\begin{equation}\label{eqn.4statepolicy}
    \widehat{\pi}_t (\text{red} \mid 2 ) = \begin{cases}
    \frac{\sum_{i=1}^{H-1} X_i U_i}{\sum_{i=1}^{H-1} X_i}, \qquad &\text{if } \sum_{i=1}^{H-1} X_i > 0,\\
    1, &\text{otherwise}.
    \end{cases}
\end{equation}

Note that this policy can be computed since at any time $t$ at which state $2$ was not visited in the dataset (i.e. $U_t$ is unknown), $X_t = 0$. Using the Poissonization trick, suppose the number of trajectories from the expert $n \sim \mathrm{Poi} (N/2)$. This is permissible since $n \le N$ with very high probability ($\ge 1 - e^{-3N/16}$ using Poisson tail bounds). Under this assumption, $X_t$'s are distributed independently as $\mathrm{Poi} \left( \frac{N}{2} \mathrm{Pr}_{\pi^*} (s_t = 2) \right)$. Using the property that for $X \sim \mathrm{Poi} (\mu)$ and independent $Y \sim \mathrm{Poi} (\lambda)$, $\mathbb{E} \left[ X/(X+Y) \mid X+Y > 0 \right] = {\mu}/(\mu + \lambda)$, we have
{\small
\begin{equation*}
    \mathbb{E} \left[ \frac{\sum_{t=1}^{H-1} X_t U_t}{\sum_{t=1}^{H-1} X_t} ~ \bigg| ~ \sum_{t=1}^H X_t > 0 \right] = \frac{\sum_{t=1}^{H-1} \mathrm{Pr}_{\pi^*} (s_t = 2) U_t}{\sum_{t=1}^{H-1} \mathrm{Pr}_{\pi^*} (s_t = 2)},
\end{equation*}}
Finally, observe that $\mathrm{Pr}_{\pi^*} (\sum_{t=1}^{H-1} X_t = 0) = \prod_{t=1}^{H-1} \mathrm{Pr}_{\pi^*} (X_t = 0) = e^{- \frac{N}{2} \sum_{t=1}^{H-1} \mathrm{Pr}_{\pi^*} (s_t = 2)}$. Therefore, 
{\small
\begin{equation*}
    \left| \mathrm{Pr}_{\pi^*} (s_H = 3) - \mathbb{E} \left[  \mathrm{Pr}_{\widehat{\pi}} (s_H = 3) \right] \right| \leq \mathrm{Pr}_{\pi^*} \left( \sum_{t=1}^{H-1} X_t = 0 \right) \sum_{t=1}^{H-1} \mathrm{Pr}_{\pi^*} (s_t = 2) \lesssim \frac{1}{N},
\end{equation*}}
since $\sup_{x} xe^{-t x} = 1/(et)$ for any $t>0$. 

We remark that~(\ref{eqn.4statepolicy}) is carefully constructed such that $\widehat{\pi}_t(\text{red}\mid 2) \in [0,1]$ almost surely to guarantee it is a valid policy, and many natural approaches such as replacing the denominator with the expectation of $\sum_{t=1}^{H-1}X_t$ does not achieve this goal.

\subsection{Mimicking a single state with no error compounding}

The proof of Theorem~\ref{thm.4statemdpunbiased} crucially relies on obtaining a policy whose expected state visitation probability at state 3 of the terminal layer is nearly the same as that of the expert. Can we generalize it to \emph{arbitrary} MDPs and \emph{arbitrary} target state? The following theorem answers this question affirmatively.

\begin{theorem}\label{theorem:single_state}
In the known-transition setting, fix any state $s^*$ at time $t$ of any MDP $\calM$. Consider a deterministic expert policy $\pi^*$, an expert dataset $D$ with $n$ trajectories, and any subset $\calS_0 \subseteq \cup_{t=1}^H\calS_t$ of states at which the expert actions are known. Let $\Pi_{\mathrm{mimic}}(\calS_0)$ be the set of policies that mimic the expert action on all states of $\calS_0$, there exists a learner $\widehat{\pi}\in \Pi_{\mathrm{mimic}}(\calS_0)$ such that 
$
|\bE[\mathrm{Pr}_{\widehat{\pi}}(s_t = s^*)] - \mathrm{Pr}_{\pi^*}(s_t = s^*)| \lesssim \frac{1}{N}. 
$
\end{theorem}
\begin{corollary}\label[corollary]{cor.3stategeneral}
Suppose $|\mathcal{S}| = 3$ and $r_t \equiv 0$ for all $t = 1,2,\ldots,H-1$, $\mathbf{r}_H \in [0,1]$, $\pi^*$ is optimal for $\mathbf{r}$, and the transitions are known. Then, there exists an efficient algorithm based on \textsc{Mimic-MD} and  \textsc{Mimic-Mixture} such that the suboptimality is upper bounded by $\widetilde{O}(1/N)$ with probability 0.99. 
\end{corollary}

The main message of Theorem \ref{theorem:single_state} is that, in the known transition setting, there is \emph{no error compounding} for achieving a near-unbiased \emph{realization} of the probability of any single state. Specifically, the upper bound $O(1/N)$ in Theorem \ref{theorem:single_state} crucially does not depend on $H$, which is in sharp contrast to the unknown transition setting where the error is $\Theta(H/N)$, as well as the known transition setting but with an absolute error $\Theta(\sqrt{H}/N)$. The construction of the policy $\widehat{\pi}$ relies on a mixture of two deterministic policies inside $\Pi_{\mathrm{mimic}}(\calS_0)$, where the choice of the mixing coefficient is much more complicated than that in the proof of Theorem~\ref{thm.4statemdpunbiased} requires a careful inductive procedure detailed later. We also note that the choice of the subset $\calS_0$ is arbitrary, and Theorem \ref{theorem:single_state} holds even if $\calS_0 = \emptyset$; the reason why we introduce $\calS_0$ is to show that the near-unbiased realization does not require a costly coordination among all states, and it could always be achieved by properly specifying the actions for a possibly small number of unvisited states.

\begin{algorithm}[htb]
	\caption{\textsc{Mimic-Mixture}}
	\label{alg:unbiased}
	\begin{algorithmic}[1]
		\State \textbf{Input:} Expert dataset $D$, states $\calS_0$ with known expert action, target state $s^*$ at time $t$
		\State Compute the following two policies $\pi^{\textrm{L}}$ and $\pi^{\textrm{S}}$ based on the known transitions: 
		\begin{equation}\label{eq:extremal_policy}
		\pi^{\textrm{L}} = \underset{\pi\in \Pi_{\mathrm{mimic}}(\calS_0)}{\textrm{argmax}} \mathrm{Pr}_{\pi}(s_t = s^*), \quad  \pi^{\textrm{S}} = \underset{\pi\in \Pi_{\mathrm{mimic}}(\calS_0)}{\textrm{argmin}} \mathrm{Pr}_{\pi}(s_t = s^*). 
		\end{equation}
		\State Draw $n\sim \Poi(N/2)$, and return an arbitrary policy $\widehat{\pi}$ if $n > N$. 
		\State For every possible trajectory $\mathsf{tr} = (s_1,\cdots,s_H) \in \calS^H$, count its number of appearances $X(\mathsf{tr})$ from the first $n$ trajectories in the expert dataset. 
		\State For each $\mathsf{tr}\in \calS^H$, compute $\beta^{\textrm{L}}(\mathsf{tr})$, $\beta^{\textrm{S}}(\mathsf{tr})$ and $\beta^*(\mathsf{tr})$ according to Lemma \ref{lemma:unbiased_coefficients}. 
		\State Subsample each $X(\mathsf{tr})$ independently with probability $\beta^{\textrm{L}}(\mathsf{tr}) - \beta^{\textrm{S}}(\mathsf{tr})$ to obtain $Y(\mathsf{tr})$. 
		\State Subsample each $Y(\mathsf{tr})$ independently with probability $(\beta^*(\mathsf{tr}) - \beta^{\textrm{S}}(\mathsf{tr}))/(\beta^{\textrm{L}}(\mathsf{tr}) - \beta^{\textrm{S}}(\mathsf{tr}))$ to obtain $Z(\mathsf{tr})$. 
		\State Compute the mixing coefficient{\small
		\begin{equation}\label{eq:mixing_coefficient}
		\widehat{\alpha} = \frac{\sum_{\mathsf{tr}\in \calS^H} Z(\mathsf{tr}) }{\sum_{\mathsf{tr}\in \calS^H} Y(\mathsf{tr})}.
		\end{equation}}
		If the denominator is zero, return any $\widehat{\alpha} \in [0,1]$.
		\State \textbf{Return} a randomized policy $\widehat{\pi} = \widehat{\alpha}\pi^{\textrm{L}} + (1-\widehat{\alpha})\pi^{\textrm{S}}$. 
	\end{algorithmic}
	\label{alg.minimcmixture}
\end{algorithm}

The construction of the learner's policy $\widehat{\pi}$ is summarized by \textsc{Mimic-Mixture} in Algorithm \ref{alg:unbiased}. The idea is to find two extremal policies, i.e. policies $\pi^{\textrm{L}}$ and $\pi^{\textrm{S}}$ which maximize and minimize the induced probability of the target state $s^*$ among all policies in $\Pi_{\mathrm{mimic}}(\calS_0)$, respectively (cf. \eqref{eq:extremal_policy}), and choose the learner's policy $\widehat{\pi}$ as a proper mixture of these extremal policies, i.e. $\widehat{\pi} = \widehat{\alpha}\pi^{\textrm{L}} + (1-\widehat{\alpha})\pi^{\textrm{S}}$. Since $\pi^{\textrm{L}}, \pi^{\textrm{S}}\in \Pi_{\mathrm{mimic}}(\calS_0)$, it is clear that the mixture $\widehat{\pi}$ also belongs to $\Pi_{\mathrm{mimic}}(\calS_0)$.  As the learner's target is to match the expert probability $\mathrm{Pr}_{\pi^*}(s_t = s^*)$, the ideal choice of $\widehat{\alpha}$ would be
\begin{align}\label{eq:mixing_coefficient_ideal}
\alpha^* = \frac{\mathrm{Pr}_{\pi^*}(s_t = s^*) - \mathrm{Pr}_{\pi^{\textrm{S}}}(s_t = s^*)}{ \mathrm{Pr}_{\pi^{\textrm{L}}}(s_t = s^*) -  \mathrm{Pr}_{\pi^{\textrm{S}}}(s_t = s^*)}, 
\end{align}
which by definition of $\pi^{\textrm{L}}, \pi^{\textrm{S}}$ always lies in $[0,1]$. Note that the only unknown quantity in \eqref{eq:mixing_coefficient_ideal} is the probability $\mathrm{Pr}_{\pi^*}(s_t = s^*)$ induced by the unknown expert policy, we need to replace this probability by a proper estimator. The most natural approach is to use the empirical version of $\mathrm{Pr}_{\pi^*}(s_t = s^*)$, which is an unbiased estimator. However, plugging this empirical version into \eqref{eq:mixing_coefficient_ideal} may make the final ratio $\alpha^*$ outside $[0,1]$, giving an invalid mixture policy $\widehat{\pi}$; a na\"{i}ve truncation of $\alpha^*$ to $[0,1]$ will also incur a too large bias (of the order $\Omega(\sqrt{H}/N)$), for the truncation operation is similar in spirit to the minimum distance projection used in \textsc{Mimic-MD}.

To circumvent this difficulty, our idea is to replace all probabilities $\mathrm{Pr}_{\pi^*}(s_t = s^*)$, $\mathrm{Pr}_{\pi^{\textrm{L}}}(s_t = s^*)$, $\mathrm{Pr}_{\pi^{\textrm{S}}}(s_t = s^*)$ in \eqref{eq:mixing_coefficient_ideal} by appropriate estimates such that the ratio lies in $[0,1]$ almost surely, even if the latter two probabilities are in fact perfectly known and thus do not require any estimation in principle. To construct these estimators, we consider a Poissonized sampling model as follows: draw an independent Poisson random variable $n\sim \Poi(N/2)$, which does not exceed $N$ with probability at least $1 - e^{-\Omega(N)}$ by the Chernoff bound. For each possible state trajectory $\mathsf{tr} = (s_1,\cdots,s_H)\in \calS^H$, define $X(\mathsf{tr})$ to be the total count of this trajectory in the first $n$ trajectories of $D$: 
\begin{align*}
X(\mathsf{tr}) = \sum_{i=1}^n \mathbbm{1} ( \mathsf{tr}_i = \mathsf{tr} ).
\end{align*}
Note that the sample size in the above count is a Poisson random variable $n\sim \Poi(N/2)$, instead of the fixed number $N$. The advantage of the Poisson sampling is that, the above count $X(\mathsf{tr})$ exactly follows a Poisson distribution $\Poi(N/2\cdot \mathrm{Pr}_{\pi^*}(\mathsf{tr}))$, and these counts $\{X(\mathsf{tr}) \}$ for different trajectories are mutually independent. We apply the following linear estimators for the probabilities in \eqref{eq:mixing_coefficient_ideal}: 
{\small
\begin{equation}\label{eq:linear_est}
\begin{split}
\widehat{\mathrm{Pr}}_{\pi^*}(s_t = s^*) &\triangleq \frac{2}{N}\sum_{\mathsf{tr}\in \calS^H} \beta^*(\mathsf{tr})\cdot X(\mathsf{tr}), \quad
\widehat{\mathrm{Pr}}_{\pi^{\textrm{L}}}(s_t = s^*) \triangleq \frac{2}{N}\sum_{\mathsf{tr}\in \calS^H} \beta^{\textrm{L}}(\mathsf{tr})\cdot X(\mathsf{tr}), \\
\widehat{\mathrm{Pr}}_{\pi^{\textrm{S}}}(s_t = s^*) &\triangleq \frac{2}{N}\sum_{\mathsf{tr}\in \calS^H} \beta^{\textrm{S}}(\mathsf{tr})\cdot X(\mathsf{tr}),
\end{split}
\end{equation}}
where $\beta^*(\mathsf{tr}), \beta^{\textrm{L}}(\mathsf{tr}), \beta^{\textrm{S}}(\mathsf{tr})\in [0,1]$ are appropriate coefficients to be specified later. We require the following three properties for these coefficients: 
\begin{enumerate}
	\item Unbiasedness: the coefficients should be chosen so that the estimators in \eqref{eq:linear_est} are unbiased in estimating the corresponding true probabilities. Mathematically, we require that
	$%\begin{align*}
	 \sum_{\mathsf{tr}\in \calS^H} \beta^\dagger(\mathsf{tr}) \cdot \textrm{Pr}_{\pi^*}(\mathsf{tr}) = \textrm{Pr}_{\pi^\dagger}(s_t = s^*), \quad \dagger \in \{*, \textrm{L}, \textrm{S} \}. 
	$%\end{align*}
	\vspace{-1.5mm}
	\item Order: for every trajectory $\mathsf{tr}\in \calS^H$, it holds that $\beta^{\textrm{S}}(\mathsf{tr})\le \beta^*(\mathsf{tr})\le \beta^{\textrm{L}}(\mathsf{tr})$. This requirement ensures that plugging \eqref{eq:linear_est} into \eqref{eq:mixing_coefficient_ideal} always gives a ratio in $[0,1]$. 
	\vspace{-1.5mm}
	\item Feasibility: this requirement is a bit subtle. We require that all coefficients $\beta^*(\mathsf{tr})$, $\beta^{\textrm{L}}(\mathsf{tr})$, and $\beta^{\textrm{S}}(\mathsf{tr})$ only depend on public information (known transition probabilities, initial distribution, expert actions at states in $\calS_0$, policies $\pi^{\textrm{L}}, \pi^{\textrm{S}}$, and $s^*$) and the private information associated with $\mathsf{tr}$ (expert actions at states visited in trajectory $\mathsf{tr}$). Importantly, these coefficients cannot depend on expert actions not in $\calS_0\cup\mathsf{tr}$, as those actions may not be observable to the learner, leaving the coefficients not always well-defined. In contrast, dependence on the expert actions at states in $\mathsf{tr}$ is feasible, for these actions are observed if $X(\mathsf{tr})>0$, and the coefficients could be arbitrarily chosen with $\beta^\dagger(\mathsf{tr})\cdot X(\mathsf{tr})\equiv 0$ if $X(\mathsf{tr})=0$, for $\dagger\in \{*,\textrm{L},\textrm{S}\}$. 
	\vspace{-1.5mm}
\end{enumerate}
The following lemma shows that we can indeed construct coefficients $\{\beta^*(\mathsf{tr})\}, \{ \beta^{\textrm{L}}(\mathsf{tr}) \}, \{ \beta^{\textrm{S}}(\mathsf{tr}) \}$ satisfying the above three requirements, and they can be used to construct a policy such that Theorem~\ref{theorem:single_state} holds.  
\begin{lemma}\label[lemma]{lemma:unbiased_coefficients}
There exist coefficients $\beta^*(\mathsf{tr}), \beta^{\textrm{\rm L}}(\mathsf{tr}), \beta^{\textrm{\rm S}}(\mathsf{tr})\in [0,1]$ such that all of the unbiasedness, order, and feasibility properties hold. Furthermore, given any such coefficients, one can efficiently construct a policy $\widehat{\pi}$ such that Theorem~\ref{theorem:single_state} holds. 
\end{lemma}
The proof of Lemma \ref{lemma:unbiased_coefficients} is via a careful inductive argument and is deferred to the Appendix. 

\section{Future work}
The main limitation of Theorem \ref{theorem:single_state} is that the construction of $\widehat{\pi}$ changes with $s^*$ and thus does not work for multiple states at the same time. Note that if one could find a policy $\widehat{\pi}$ such that the inequality in Theorem \ref{theorem:single_state} holds for all states simultaneously, then it is clear that this policy would achieve an expected suboptimality at most $O(|\calS|H/N)$ in imitation learning, which would crucially imply no error compounding in imitation learning with known transition. It is an outstanding open question to construct such a policy $\widehat{\pi}$ or establish the impossibility, and we leave it for future work. 

\appendix

%% file: Appendix-A.tex
\section{}

\subsection{Proof of Theorem~\ref{thm.reductions}}

\begin{proof}
(IL $\longrightarrow$ value estimation): for the case of reward function being symmetric, by choosing both $\mathbf{r}$ and $1-\mathbf{r}$ and union bound we have the desired result. For the expert optimal case, we can save one $\delta$ factor since we know $|J_{\mathbf{r}}(\pi^*) - J_{\mathbf{r}}(\widehat{\pi})| = J_{\mathbf{r}}(\pi^*) - J_{\mathbf{r}}(\widehat{\pi})$.

(value estimation $\longrightarrow$ IL): To analyze the suboptimality of this learner, observe that,
\begin{align}
    J_{\mathbf{r}} (\pi^*) - J_{\mathbf{r}}(\widehat{\pi}) &\le \max_{\mathbf{r}' \in \mathcal{R}_D}  J_{\mathbf{r}'} (\pi^*) - \widetilde{J}_{\mathbf{r}'} (\pi^*)  + \max_{\mathbf{r}' \in \mathcal{R}_D}  \widetilde{J}_{\mathbf{r}'} (\pi^*) - J_{\mathbf{r}'} (\widehat{\pi})  \\
    &\overset{(i)}{\le} \max_{\mathbf{r}' \in \mathcal{R}_D} | J_{\mathbf{r}'} (\pi^*) - \widetilde{J}_{\mathbf{r}'} (\pi^*) | + \max_{\mathbf{r}' \in \mathcal{R}_D}  \widetilde{J}_{\mathbf{r}'} (\pi^*) - J_{\mathbf{r}'} (\pi^*)  \le 2 \epsilon.
\end{align}
where $(i)$ uses the fact that $\pi^*$ is a feasible policy to the optimization problem \eqref{eq:value-learner}.
\end{proof}

\subsection{Proof of Theorem~\ref{thm.mimimdefficient}}\label{sec.mimicefficient}

\begin{algorithm}[t]
	\caption{\textsc{Mimic-MD}}
	\label{alg:det:NsimInf}
	\begin{algorithmic}[1]
		\State \textbf{Input:} Expert dataset $D$.
		\State Choose a uniformly random permutation of $D$,
		\Statex Define $D_1$ to be the first $N/2$ trajectories of $D$ and $D_2 = D \setminus D_1$.
		\State Define $\mathcal{T}^{D_1}_t ( s,a) \triangleq \{ \{ (s_{t'},a_{t'}) \}_{t'=1}^H | s_t{=}s, a_t{=}a,\ \exists \tau \le t : s_\tau \not\in \mathcal{S}_\tau (D_1) \}$ as trajectories that visit $(s,a)$ at time $t$, and at some time $\tau {\le} t$ visit a state unvisited at time $\tau$ in any trajectory in $D_1$.
		\State Define the optimization problem \eqref{eq:opt} below and return $\widehat{\pi}$ as any optimizer of it:
		\begin{equation} \label{eq:opt}
		\min_{\pi \in \Pi_{\mathrm{mimic}} (D_1)} \ \sum_{t=1}^H \sum_{(s,a) \in \mathcal{S} \times \mathcal{A}} \left| \mathrm{Pr}_\pi \Big[ \mathcal{T}^{D_1}_t ( s,a) \Big] - \frac{1}{|D_2|}\sum_{\textsf{tr} \in D_2} \mathbbm{1} \left( \textsf{tr} \in \mathcal{T}^{D_1}_t ( s,a) \right) \right|. \tag{\textsf{OPT-MD}}
		\end{equation}
		\Statex \Comment{$\Pi_{\mathrm{mimic}} (D_1)$ is the set of policies that mimics the expert on the states visited in $D_1$ (\cref{eq:Pi.mimic})}
		\State \textbf{Return} $\widehat{\pi}$
	\end{algorithmic}
\end{algorithm}

We show that solving the objective \eqref{eq:opt} in \textsc{Mimic-MD} can be posed as a convex program and is thus computationally tractable. Specifically, the family of the learner's policy $\pi$ could be represented by a set of joint state-action probabilities $\{p_t^\pi(s_t,a_t)\}_{t\in [H], s_t\in \calS, a_t\in \calA}\in \Omega$, where $\Omega$ is the feasible set of all possible $\{q_t(s_t,a_t)\}$ such that the following constraints hold: 
\begin{align}
&\sum_{s_t\in \calS}  \sum_{a_t\in \calA} q_t(s_t,a_t)P_t(s_{t+1}\mid s_t,a_t) = \sum_{a_{t+1}\in \calA} q_{t+1}(s_{t+1},a_{t+1}), \qquad \forall t\in [H-1], s_{t+1}\in \calS; \label{eq:cnstr_compatible} \\
&\sum_{a_1\in \calA} q_1(s_1,a_1) = \rho(s_1), \qquad \forall s_1\in \calS; \label{eq:cnstr_initial} \\
&q_t(s_t,a_t) = 0, \qquad \forall t\in [H], s_t\in \calS_t(D_1) \text{ and } a_t\neq \pi^*(s_t); \label{eq:cnstr_mimic} \\
&q_t(s_t,a_t) \ge 0, \qquad \forall t\in [H], s_t \in \calS, a_t \in \calA. \label{eq:cnstr_prob}
\end{align}
Given a feasible solution $\{q_t(s_t,a_t)\}$ satisfying \eqref{eq:cnstr_compatible} to \eqref{eq:cnstr_prob}, the (randomized) learner's policy $\widehat{\pi}$ is constructed via $\mathrm{Pr}(\widehat{\pi}(s_t) = a_t) = q_t(s_t,a_t) / \sum_{\tilde{a}_t \in \calA} q_t(s_t,\tilde{a}_t)$. We prove that $\widehat{\pi}\in \Pi_{\mathrm{mimic}}(D_1)$ and $q_t(s,a) = \mathrm{Pr}_{\widehat{\pi}}[s_t=s,a_t=a]$ for $t\in [H], s\in \calS, a\in \calA$, thereby establish a one-to-one correspondance between all feasible policies $\Pi_{\mathrm{mimic}}(D_1)$ and the feasible set $\Omega$. First, the non-negativity constraint \eqref{eq:cnstr_prob} implies that $\widehat{\pi}$ is a valid randomized policy, and \eqref{eq:cnstr_mimic} shows that $\widehat{\pi}$ mimics the expert policy on $D_1$, i.e. $\widehat{\pi}\in \Pi_{\mathrm{mimic}}(D_1)$. Second, for $t=1$, the identity \eqref{eq:cnstr_initial} implies that
\begin{align*}
\mathrm{Pr}_{\widehat{\pi}}[s_1 = s, a_1 = a] = \rho(s_1)\cdot \frac{q_1(s,a)}{\sum_{\tilde{a}\in\calA} q_1(s,\tilde{a})} = \rho(s_1)\cdot \frac{q_1(s,a)}{\rho(s_1)} = q_1(s,a), 
\end{align*}
as claimed. Finally, suppose that $q_{t}(s,a) = \mathrm{Pr}_{\widehat{\pi}}[s_{t}=s,a_{t}=a]$ holds for some $t\in [H-1]$, then for time $t+1$, the compatibility condition \eqref{eq:cnstr_compatible} gives that 
\begin{align*}
\mathrm{Pr}_{\widehat{\pi}}[s_{t+1} = s, a_{t+1} = a] &= \mathrm{Pr}_{\widehat{\pi}}[s_{t+1} = s] \cdot \frac{q_{t+1}(s,a)}{\sum_{\tilde{a}} q_{t+1}(s,\tilde{a})} \\
&= \left(\sum_{s'\in \calS} \sum_{a'\in \calA} \mathrm{Pr}_{\widehat{\pi}}[s_{t} = s', a_{t} = a'] \cdot P_t(s\mid s',a') \right) \cdot \frac{q_{t+1}(s,a)}{\sum_{\tilde{a}\in\calA} q_{t+1}(s,\tilde{a})} \\
&= \left(\sum_{s_t\in \calS} \sum_{a_t\in \calA} q_t(s_t,a_t) \cdot P_t(s\mid s_t,a_t) \right) \cdot \frac{q_{t+1}(s,a)}{\sum_{\tilde{a}\in\calA} q_{t+1}(s,\tilde{a})} \\
&= q_{t+1}(s,a). 
\end{align*}
Therefore by induction, we conclude that any element of the feasible set $\Omega$ gives rise to a feasible policy $\widehat{\pi}\in \Pi_{\mathrm{mimic}}(D_1)$, and the reversed direction is obvious. Consequently, given the feasible set $\Omega$, the \textsc{Mimic-MD} objective \eqref{eq:opt} solves the following linear program:
\begin{equation}\label{eq:mimic-md_LP}
\begin{split}
\text{minimize} &\qquad \sum_{t=1}^H \sum_{(s,a)\in \calS\times \calA} \left| q_t(s,a) - \frac{1}{|D_2|}\sum_{\mathsf{tr} \in D_2} \mathbbm{1} \left(\mathsf{tr}(s_t,a_t) = (s,a) \right) \right|, \\
\text{subject to}&\qquad \{q_t(s_t,a_t)\}_{t\in [H], s\in \calS, a\in \calA}\in \Omega.
\end{split}
\end{equation}
It is clear that the linear program \eqref{eq:mimic-md_LP} has $O(|\calS||\calA|H)$ variables and $O(|\calS||\calA|H)$ linear constraints, and therefore \textsc{Mimic-MD} can be solved in $\textrm{poly}(|\calS|,|\calA|,H)$ time. 

Now we show the upper bound when $|\mathcal{S}| = 2$. The case for $|\mathcal{S}|\geq 3$ can be found in~\citep{rajaraman2020fundamental}. 

Let $E_t$ be the event that there exists one state at time $t$ that has not been visited in the $N$ expert trajectories. Note that we know the policy $\pi^*_t$ exactly if both states at time $t$ have been visited: it implies that given an estimator $\widehat{f}_t^{\pi^*}$ for $f_t^{\pi^*}$, if we have seen both states for $t' = t+1,t+2,\ldots,t+m$, we can estimate $f_{t'}^{\pi^*}$ via computing the marginal distributions at these time steps $t'$ since we know the conditional distributions exactly. Using the data processing inequality (Lemma~\ref{lemma:TVshrink}) below, we know that $\mathsf{TV}(f_{t'}^{\pi^*},\widehat{f}_{t'}^{\pi^*})\leq \mathsf{TV}(f_t^{\pi^*},\widehat{f}_t^{\pi^*})$. Hence, we have 
\begin{align}
        \sum_{t=1}^H \mathsf{TV}(f_t^{\pi^*},\widehat{f}_t^{\pi^*}) \leq H \max_{t: E_t\text{ holds}} \mathsf{TV}(f_t^{\pi^*},\widehat{f}_t^{\pi^*}).
\end{align}
It follows from the Binomial distribution formula that the marginal probability for the unseen states for each $E_i$ is at most $\log(H/\delta)/N$ with probability at least $1-\delta/H$ for each $i$. By union bound we know that with probability at least $1-\delta$, for all time steps $t$ such that $E_t$ is true, the unseen state has marginal probability $\lesssim \log(H/\delta)/N$. In other words, with high probability, the state distribution at each time $t$ with an unobserved is of the form $(p,1-p)$ where $p \lesssim \log(H/\delta)/N$. For such a distribution, using \cite[Lemma 4]{han2015minimax}, the empirical distribution achieves \textsf{TV} error $\lesssim \sqrt{\frac{p}{N}} \lesssim \sqrt{\log(H/\delta)}/N$ for each $t$, which results in the final $H\times \sqrt{\log(H/\delta)}/N$ bound on $\sum_{t=1}^H \mathsf{TV}(f_t^{\pi^*},\widehat{f}_t^{\pi^*})$. Finally, observe that the suboptimality of \textsc{Mimic-MD} is upper bounded by this quantity, since $J(\pi^*) - J(\widehat{\pi}) = \sum_{t=1}^H \mathbb{E}_{\pi^*} \left[ \mathbf{r}_t (s_t,a_t) \right] - \mathbb{E}_{\widehat{\pi}} \left[ \mathbf{r}_t (s_t,a_t) \right] = \sum_{t=1}^H \textsf{TV} (f_{\pi^*}, f_{\widehat{\pi}})$.
The expected suboptimality bound directly follows from integrating the high probability bound using $\mathbb{E}[X] = \int_{0}^\infty \mathbb{P}(X > t) dt$ for nonnegative random variables. 

\begin{lemma} \label[lemma]{lemma:TVshrink}
Consider any distributions $p,q$ supported on $[n]$. Let $P$ be any Markov kernel from $[n] \to \Delta_{[n]}$. Then $\textsf{TV} (P \circ p, P \circ q) \le \textsf{TV} (p,q)$.
\end{lemma}
\begin{proof}
$\sum_{j=1}^n \left| \sum_{i \in n} p_i P_{ij} - \sum_{i \in n} q_i P_{ij} \right| = \sum_{j=1}^n \left| \sum_{i \in n} (p_i - q_i) P_{ij} \right| \le \sum_{j=1}^n \sum_{i \in n} |p_i - q_i| P_{ij} = \sum_{i=1}^n |p_i - q_i|$.
\end{proof}

%------------------------------------------------------------------------
%------------------------------------------------------------------------
%------------------------------------------------------------------------
%------------------------------------------------------------------------

\subsection{Proof of Theorem~\ref{theorem:lb-known-transition}} \label{sec.lowerbound-knowntr}

The first key observation is first that in order to establish a lower bound on the one-sided error probability $\mathrm{Pr} (J_{\mathcal{M}} (\pi^*) - J_{\mathcal{M}} (\widehat{\pi}) \ge H^{3/2} / N )$ for any learner $\widehat{\pi}$, it suffices to lower bound the two-sided error probability $\mathrm{Pr} ( \left| J_{\mathcal{M}} (\pi^*) - J_{\mathcal{M}} (\widehat{\pi}) \right| \ge H^{3/2} / N )$. Intuitively this is because the learner gets no reward feedback - a learner which has small one-sided error probability on some MDP $\mathcal{M} = (P,\mathbf{r})$ can potentially have large one-sided error probability on the MDP $\mathcal{M} = (P, 1-\mathbf{r})$. In the absence of reward feedback, the learner cannot distinguish between these two cases. The only option for the learner is to guarantee small two-sided error probability on all IL instances to ensure a uniform bound on the one-sided error probability. In particular, we show the following result.

\begin{lemma} \label[lemma]{lemma:2sided-1sided}
Suppose there exists an MDP $\mathcal{M}$ with $|\mathcal{S}| = 3$ such that,
$\mathrm{Pr} \left( \left| J_{\mathcal{M}} (\pi^*) - J_{\mathcal{M}} (\widehat{\pi} (D)) \right| \gtrsim \frac{H^{3/2}}{N} \right) \ge c'$ for some constant $0 < c' \le 1$. Then there exists an MDP $\mathcal{M}'$ with $|\mathcal{S}| = 3$ such that, $\mathrm{Pr} \left( J_{\mathcal{M}'} (\pi^*) - J_{\mathcal{M}'} (\widehat{\pi} (D)) \gtrsim \frac{H^{3/2}}{N} \right) \ge c'/2$.
\end{lemma}
\begin{proof}
Suppose for every MDP $\mathcal{M}$, there exists a learner $\widehat{\pi}$ such that,
\begin{equation} \label{eq:dev1}
    \mathrm{Pr} \left( J_{\mathcal{M}} (\pi^*) - J_{\mathcal{M}} (\widehat{\pi} (D)) \gtrsim \frac{H^{3/2}}{N} \right) < \frac{c'}{2}.
\end{equation}
This implies that for every MDP $\mathcal{M}$,
\begin{equation} \label{eq:dev2}
    \mathrm{Pr} \left( J_{\mathcal{M}} (\widehat{\pi} (D)) - J_{\mathcal{M}} (\pi^*) \gtrsim \frac{H^{3/2}}{N} \right) < \frac{c'}{2}.
\end{equation}
This follows from the fact that for any MDP $\mathcal{M} = (P,\mathbf{r})$ we can consider an MDP $\mathcal{M}' = (P,\mathbf{r}')$ where $\mathbf{r}_t' = 1 - \mathbf{r}_t$. As a consequence $J_{\mathcal{M}'} (\pi) = H - J_{\mathcal{M}} (\pi)$ for any policy $\pi$ which gives the equation. By adding together \cref{eq:dev1,eq:dev2} we see that for every MDP $\mathcal{M}$, $\widehat{\pi}$ satsifies the property that
\begin{equation}
    \mathrm{Pr} \left( \left| J_{\mathcal{M}} (\widehat{\pi} (D)) - J_{\mathcal{M}} (\pi^*) \right| \gtrsim \frac{H^{3/2}}{N} \right) < c'.
\end{equation}
Taking the contrapositive, this implies the required statement.
\end{proof}

In order to furnish the lower bound, we will consider a Bayes IL problem, where the expert's policy $\pi^*$ and the underlying MDP are sampled from some distribution $\mathcal{D}$.

In order to prove this result, we assume that the underlying MDP $\mathcal{M}$ and the expert policy $\pi^*$ are jointly sampled from a distribution $\mathcal{D}$ and show that there is a constant $c'$ such that
\begin{equation}
    \mathbb{E}_{(\pi^*, \mathcal{M}) \sim \mathcal{D}} \left[ \mathbb{E} \left[ \mathbbm{1} \left( \left| J_{\mathcal{M}} (\pi^*) - J_{\mathcal{M}} (\widehat{\pi}) \right| \lesssim \frac{H^{3/2}}{N} \right) \right] \right] < c'.
\end{equation}
This implies the existence of an MDP $\mathcal{M}$ and expert policy $\pi^*$ with the required property. Next, we use a symmetrization argument to upper bound the LHS of the above formula.

\begin{lemma} \label[lemma]{lemma:symmetrization}
For any constant $C > 0$,
\begin{equation}
    \mathrm{Pr} \left( \left| J_{\mathcal{M}} (\pi^*) - J_{\mathcal{M}} (\widehat{\pi}) \right| \le \frac{C H^{3/2}}{N} \right) \le \frac{1}{2} + \frac{1}{2}\mathbb{E} \left[ \mathrm{Pr} \left( \left| J_{\mathcal{M}} (\pi^*_1) - J_{\mathcal{M}} (\pi_2^*) \right| \le \frac{C H^{3/2}}{N} \middle| \mathcal{M}, D \right) \right]
\end{equation}
where $\pi_1^*$ and $\pi^*_2$ are independent copies of the expert's policy drawn from the posterior distribution conditioned on the expert dataset $D$ and MDP $\mathcal{M}$.
\end{lemma}
\begin{proof}
By definition,
\begin{align}
    &2 \mathrm{Pr} \left( \left| J_{\mathcal{M}} (\pi^*) - J_{\mathcal{M}} (\widehat{\pi}) \right| \le \frac{C  H^{3/2}}{N} \right) \\
    &\overset{(i)}{=} \mathbb{E} \left[ \mathrm{Pr} \left( \left| J_{\mathcal{M}} (\pi^*_1) - J_{\mathcal{M}} (\widehat{\pi}) \right| \le \frac{C  H^{3/2}}{N} \middle| \mathcal{M}, \widehat{\pi} \right) \right] + \mathbb{E} \left[ \mathrm{Pr} \left( \left| J_{\mathcal{M}} (\pi^*_2) - J_{\mathcal{M}} (\widehat{\pi}) \right| \le \frac{C  H^{3/2}}{N} \middle| \mathcal{M}, \widehat{\pi} \right) \right] \\
    &\overset{(ii)}{\le} 1+\mathbb{E} \left[ \mathrm{Pr} \left( \left| J_{\mathcal{M}} (\pi^*_1) - J_{\mathcal{M}} (\widehat{\pi}) \right| + \left| J_{\mathcal{M}} (\pi^*_2) - J_{\mathcal{M}} (\widehat{\pi}) \right|  \le \frac{C  H^{3/2}}{N} \middle| \mathcal{M}, \widehat{\pi} \right) \right] \\
    &\overset{(iii)}{\le} 1+\mathbb{E} \left[ \mathrm{Pr} \left( \left| J_{\mathcal{M}} (\pi^*_1) - J_{\mathcal{M}} (\pi^*_2) \right| \le \frac{C  H^{3/2}}{N} \middle| \mathcal{M}, \widehat{\pi} \right) \right] \\
    &= 1+\mathbb{E} \left[ \mathrm{Pr} \left( \left| J_{\mathcal{M}} (\pi^*_1) - J_{\mathcal{M}} (\pi^*_2) \right| \le \frac{C  H^{3/2}}{N} \middle| \mathcal{M}, D \right) \right]
\end{align}
where in $(i)$, $\pi_1^*$ and $\pi^*_2$ are as defined in the theorem statement. $(ii)$ uses the fact that $\mathbbm{1} (x \le a) + \mathbbm{1} (y \le b) \le 1+\mathbbm{1} (x + y \le a+b)$ and $(iii)$ follows by triangle inequality. The last inequality follows from the fact that the expert policy is independent of any external randomness employed by $\widehat{\pi}$.
\end{proof}

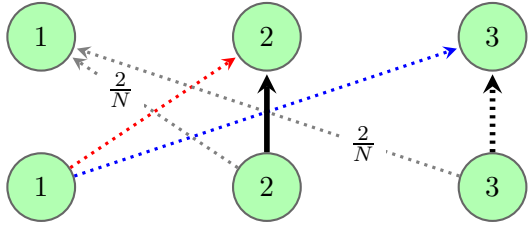
\begin{figure}
\centering
\begin{tikzpicture}[shorten >=1pt,node distance=1.25cm,on grid,auto,good/.style={circle, draw=black!60, fill=green!30!white, thick, minimum size=9mm, inner sep=0pt},bad/.style={circle, draw=black!60, fill=red!30, thick, minimum size=9mm, inner sep=0pt}]
\tikzset{every edge/.style={very thick, draw=black}}
    
\node[good]     (s1)                        {$1$};
\node[good]     (s2) [right = 3cm of s1]    {$2$};
\node[good]     (s3) [right = 3cm of s2]    {$3$};

\node[good]     (s1new) [above = 2cm of s1]  {$1$};
\node[good]     (s2new) [right = 3cm of s1new]  {$2$};
\node[good]     (s3new) [right = 3cm of s2new]  {$3$};

\path[->,>=stealth]
(s1)    edge [dotted, red] node {} (s2new)
(s3)    edge [dotted, line width=0.75mm] (s3new)
(s2)    edge [gray, dotted] node [black, anchor=center, pos=0.7, fill=white] {$\frac{2}{N}$}  (s1new)
(s3)    edge [gray, dotted] node [black, anchor=center, pos=0.25, fill=white] {$\frac{2}{N}$} (s1new)
(s1)    edge [dotted, blue] (s3new)
(s2)    edge [line width=0.75mm] (s2new);
\end{tikzpicture}
\caption{Lower bound instance for $|\mathcal{S}| = 3$. The dotted transitions offer no reward and solid transitions offer reward $1$. State $1$ is the only one with $2$ actions: red leading to state $2$ and blue leading to state $3$. The action at state $2$ and $3$ transitions the learner to state $1$ with probability $\frac{1}{N}$ and leaves it unchanged otherwise. The initial distribution is at state $2$ with probability $1$}
\label{fig:3-state-MDP}
\end{figure}

\paragraph{Lower bound instance for known transition tabular setting}

In this section, we describe the prior distribution $\mathcal{D}$ jointly over expert policies and MDPs. The MDP is time invariant. We first describe the transition structure of the MDP. We assume that $N \ge |\mathcal{S}| H$ and $|\mathcal{S}| \ge 3$. We round $|\mathcal{S}|$ down to the nearest multiple of $3$ (by making the remaining states dummy) and partition the MDP into $|\mathcal{S}|/3$ groups of $3$ states each.

\paragraph{MDP transition structure:} We prove the lower bound for the case of $|\mathcal{S}| = 3$ and defer the upper bound later. The transition of the MDP is depicted in \cref{fig:3-state-MDP}. The initial distribution of the MDP is at state $2$ with probability $1$. We assume that $|\mathcal{A}|=2$ and furthermore that the states $2$ and $3$ only have a single action. This is without loss of generality, by assuming that the two actions induce the same distribution over states and constraining the reward to be the same.

At states $2$ and $3$, playing either action transitions the learner to state $1$ with probability $1/N$ and stays put with the remaining probability. On the other hand, at state $1$, picking action $a_1$ deterministically transitions the learner to state $2$ while picking actions $a_2$ transitions the learner to state $3$. State $1$ is the only one where the choice of action is relevant so we specify a policy by only mentioning the action distribution at state $1$ in each group at each time in the episode.

\paragraph{MDP reward structure:} In each group $g$, the reward function of the MDP is chosen to be $1$ for the action at state $g_2$ and $0$ for every other state-action combination.

\paragraph{Expert policy:} The expert policy is time variant. Recall that state $g_1$ in each group $g$ is the only state where the choice of action plays a non-trivial role. Define $\Pi_{\mathrm{det}}$ as the set of all time-variant deterministic policies. Recall the assumption that in each group $g$, states $g_2$ and $g_3$ have only a single action.

To finally obtain the lower bound, we simply invoke the symmetrization argument in \Cref{lemma:symmetrization}. First, conditioned on the dataset $D$, we compute the posterior distribution of the expert policy. To this end, define $\Pi_{\mathrm{mimic}} (D)$ as the set of deterministic policies which are ``consistent'' with the dataset $D$ and at each state visited in $D$ play the same action as observed in $D$. In other words,
\begin{equation} \label{eq:Pi.mimic}
    \Pi_{\mathrm{mimic}} (D) \triangleq \Big\{ \pi \in \Pi_{\mathrm{det}} : \forall t \in [H],\ s \in \mathcal{S}_t (D),\ \pi_t (\cdot | s) = \delta_{\pi^*_t (s)} \Big\},
\end{equation}
where $\mathcal{S}_t (D)$ denotes the set of states visited at time $t$ in some trajectory in $D$, and $\pi_t^* (s)$ is the unique action played by the expert at time $t$ in any trajectory in $D$ that visits the state $s$ at time $t$. Invoking \cite[Lemma A.14]{rajaraman2020fundamental}, it follows that:

\begin{lemma} \label[lemma]{lemma:conditional-expert}
Conditioned on the expert dataset $D$, the expert policy is distributed as $\mathrm{Unif} (\Pi_{\mathrm{mimic}} (D))$. In other words, at each time $t$ such that state $1$ is unvisited in any trajectory in the expert dataset, $\pi_t^* (a_1 | 1) \sim \mathrm{Unif} (\{ 0,1\})$.
\end{lemma}

Finally, consider $\pi^*_t ( a_1 | 1)$, which is an indicator random variable for the event that the expert plays action $a_1$ at the state $1$ at time $t$. With this notation, we can compute the total reward collected by the expert policy.

\begin{lemma} \label[lemma]{lemma:total-reward}
Consider the expert policy $\pi^*$. Then,
\begin{equation}
    J_{\mathcal{M}} (\pi^*) = \sum_{t=1}^{H-1} \left( \sum_{t' = t+1}^H \left( 1 - \frac{1}{N} \right)^{H-t'} \right) \mathrm{Pr} (s_t = 1) \pi_t^* (a_1 | g_1) + \sum_{t=1}^H \left( 1 - \frac{2}{N} \right)^{t-1}.
\end{equation}
\end{lemma}
\begin{proof}
Fixing the expert policy $\pi^*$, the probability that the expert visits the state $1$ at time $2$ satisfies the condition:
\begin{align}
    \mathrm{Pr}_{\pi^*} (s_t = 1) &= \frac{2}{N} \left( \mathrm{Pr}_{\pi^*} (s_{t-1} = 2) + \mathrm{Pr}_{\pi^*} (s_{t-1} = 3) \right) \\
    \implies \mathrm{Pr}_{\pi^*} (s_t = 1) &= \frac{2}{N} (1 - \mathrm{Pr} (s_{t-1} = 1)).
\end{align}
With the initial condition $\mathrm{Pr}_{\pi^*} (s_1 = 1) = 0$, the solution to the recurrence relation is, $\mathrm{Pr}_{\pi^*} (s_t = 1) = \frac{1}{(N/2)+1} \left( 1 - \frac{1}{(-N/2)^{t-1}} \right)$. Note that this probability is independent of the actions chosen by the expert at state $2$ so henceforth we denote it by $\mathrm{Pr} (s_t = 1)$. Moreover for $t > 1$,
\begin{equation} \label{eq:P-bound-st=1}
    \frac{2(N-2)}{N^2} \le \mathrm{Pr} (s_t = 1) \le \frac{2}{N}
\end{equation}
with the upper bound for $t = 2$ and the lower bound for $t=3$. Next observe that,
\begin{align} \label{eq:recursion}
    \mathrm{Pr}_{\pi^*} (s_t = 2) = \left( 1 - \frac{2}{N} \right) \mathrm{Pr}_{\pi^*} (s_{t-1} = 2) + \mathrm{Pr} \left( s_{t-1} = 1 \right) \pi^*_t (a_1 | 1)
\end{align}
observe that $\pi^*_t (a_1|1)$ is a $\mathsf{Bern} (1/2)$ random variable indicating whether the expert picks the action $a_1$ at state $1$ at time $t$. Finally,
\begin{align}
    J (\pi^*) = \sum_{t=1}^H \mathrm{Pr}_{\pi^*} (s_t = 2) = \sum_{t=1}^{H-1} \left( \sum_{t' = t+1}^H \left( 1 - \frac{2}{N} \right)^{H-t'} \right) \mathrm{Pr} (s_t = 1) \pi_t^* (a_1 | 1)  + \sum_{t=1}^H \left( 1 - \frac{2}{N} \right)^{t-1}.
\end{align}
where the last equation uses the recursion for $\mathrm{Pr}_{\pi^*} (s_t = 2)$ in \cref{eq:recursion}.
\end{proof}

\begin{lemma} \label[lemma]{lemma:exp-value-diff}
Conditioned on the expert dataset $D$, sample two instances of the expert policy $\pi_1^*$ and $\pi^*_2$. Then,
\begin{equation}
    J_{\mathcal{M}} (\pi^*_1) - J_{\mathcal{M}} (\pi^*_2) = \sum_{t=1}^{H-1} \left( \sum_{t' = t+1}^H \left( 1 - \frac{2}{N} \right)^{H-t'} \right) \mathrm{Pr} (s_t = 1) X_t \mathbbm{1} (1 \in \mathcal{S}_t (D)).
\end{equation}
where recall that $\mathcal{S}_t (D)$ is the set of states visited in some trajectory in the dataset $D$, and $X_t$ are i.i.d. random variables distributed as:
\begin{equation}
    X_t (i) = \begin{cases}
    -1, &\quad \text{w.p. } \frac{1}{4} \\
    0, &\quad \text{w.p. } \frac{1}{2} \\
    +1, &\quad \text{w.p. } \frac{1}{4}
    \end{cases}
\end{equation}
\end{lemma}
\begin{proof}
Invoking \Cref{lemma:conditional-expert,lemma:total-reward} for $\pi_1^*$ and $\pi_2^*$, the statement follows immediately.
\end{proof}

\begin{lemma} \label[lemma]{lemma:final-exp-value-diff-bound}
There exists a constant $C > 0$ such that, if $N \ge \max \{ 7, H\}$,
\begin{align}
    &\mathbb{E} \left[ \mathrm{Pr} \left( \left| J_{\mathcal{M}} (\pi^*_1) - J_{\mathcal{M}} (\pi^*_2) \right| \le \frac{C H^{3/2}}{N} \middle| D \right) \right] \le 0.9.
\end{align}
\end{lemma}
\begin{proof}
Define the zero-mean random variable, $Z_D = J(\pi_1^*) - J(\pi^*_2)$ where $\pi_1^*$ and $\pi_2^*$ are sampled from the posterior distribution conditioned on the expert dataset $D$. From \Cref{lemma:exp-value-diff}, observe that $Z_D = \sum_{t=1}^{H-1} \kappa_t X_t$ where $\kappa_t = \sum_{t' = t+1}^H \left( 1 - \frac{2}{N} \right)^{H-t'} \mathrm{Pr} (s_t = 1) \mathbbm{1} (1 \in \mathcal{S}_t (D))$.  
By the Paley Zygmund inequality, for $0 \le \theta \le 1$,
\begin{align}
    \mathrm{Pr} \left( Z_D^2 \ge \theta \mathrm{Var} (Z_D) \middle| D \right) \ge (1 - \theta)^2 \frac{\mathbb{E} \left[ Z_D^2 \middle| D \right]^2}{\mathbb{E} \left[ Z_D^4 \middle| D \right]}.
\end{align}
Then, $\mathrm{Var} (Z_D) = \mathbb{E} \left[ Z_D^2 \middle| D \right] = \frac12 \sum_{t=1}^H \kappa_t^2$. Furthermore, $\mathbb{E} [Z^4_D] \le \frac34 \sum_{t_1 \ne t_2 \in [H]} \kappa^2_{t_1} \kappa_{t_2}^2 + \frac12 \sum_{t=1}^H \kappa_t^4 \le \frac{3}{4} (\sum_{t=1}^H \kappa_t^2 )^2$. Therefore, with $\theta = \frac{1}{10}$,
\begin{equation} \label{eq:ZD-bound}
    \mathrm{Pr} \left( Z_D^2 \ge \frac{1}{10} \mathbb{E} \left[ Z_D^2 \middle| D \right] \right) \ge \frac{99}{100} \frac{1/4}{3/4} = \frac{33}{100}.
\end{equation}

\begin{lemma} \label[lemma]{lemma:LB-E[ZD2]}
$\mathbb{E} [Z_D^2] \gtrsim \frac{H^3}{N^2}$.
\end{lemma}
\begin{proof}
By definition, $\mathbb{E} \left[ Z_D^2 \middle| D \right] = \frac12 \sum_{t=1}^H \kappa_t^2 = \frac12 \sum_{t=1}^H \left( \sum_{t' = t+1}^H \left( 1 - \frac{2}{N} \right)^{H-t'} \right)^2 \left( \mathrm{Pr} (s_t = 1) \right)^2 \mathbbm{1} (1 \in \mathcal{S}_t (D))$. Then,
\begin{align}
    \mathbb{E} \left[ Z_D^2 \right] &= \frac12 \sum_{t=1}^{H-1} \left( \sum_{t' = t+1}^H \left( 1 - \frac{2}{N} \right)^{H-t'} \right)^2 \left( \mathrm{Pr} (s_t = 1) \right)^2 \mathrm{Pr} (1 \in \mathcal{S}_t (D)) \\
    &\overset{(i)}{\gtrsim} \sum_{t=1}^{H-1} \left( H-t \right)^2 \left( \mathrm{Pr} (s_t = 1) \right)^2 \mathrm{Pr} (1 \in \mathcal{S}_t (D)) \\
    &\overset{(ii)}{\gtrsim} \frac{H^3}{N^2}, \label{eq:exp-LB}
\end{align}
Note that $(i)$ follows from the fact that $\sum_{t'=t+1}^H \left( 1 - \frac{1}{N} \right)^{H-t'} \gtrsim H-t$ since $N \ge |\mathcal{S}| H$, while $(ii)$ follows from \Cref{eq:P-bound-st=1} which shows that $\mathrm{Pr} (s_t = 1) \gtrsim \frac{1}{N}$ and the fact that $\mathrm{Pr} (1 \in \mathcal{S}_t (D)) = 1 - \left( 1 - \mathrm{Pr} (s_t = 1) \right)^N \ge 1 - \left( 1 - \frac{2(N-2)}{N^2} \right)^N \ge 4/5$ for $N \ge 7$.
\end{proof}

Next observe that,
\begin{align}
    \sqrt{\mathrm{Var} \left( \mathbb{E} \left[ Z_D^2 \middle| D \right] \right)} &\overset{(i)}{\le} \frac12 \sum_{t=1}^{H-1} \sqrt{\mathrm{Var} (\kappa_t^2)} \\
    &\le \frac12 \sum_{t=1}^{H-1} \sqrt{\mathbb{E} \left[ \kappa_t^4 \right]} \\
    &\overset{(ii)}{\le} \frac12 \sum_{t=1}^{H-1} \frac{\mathbb{E} [\kappa_t^2]}{\sqrt{4/5}} \le \frac{\mathbb{E} [Z_D^2]}{\sqrt{4/5}}. \label{eq:var-UB}
\end{align}
In $(i)$, we use the definition $\mathbb{E} \left[ Z_D^2 \middle| D \right] = \frac{1}{2} \sum_{t=1}^{H-1} \kappa_t^2$. In $(ii)$, we use the fact that $\mathbb{E} [ \kappa_t^4 ]$ is a scaled indicator random variable. Therefore, $\mathbb{E} [ \kappa_t^4 ] = \frac{\mathbb{E} [\kappa_t^2]^2}{\mathrm{Pr} (\kappa_t > 0)} \le \frac{\mathbb{E} [\kappa_t^2]^2}{4/5}$. Here, the last inequality uses the fact that $\mathrm{Pr} (\kappa_t > 0) = \mathrm{Pr} (1 \in \mathcal{S}_t (D)) \ge 1 - (1 - \mathrm{Pr} (s_t = 1))^N \ge 1 - \left(1 - \frac{2(N-1)}{N^2} \right)^N \ge 4/5$ for $N \ge 7$. Finally, by an application of the second moment method,
\begin{align}
    \mathrm{Pr} \left( \mathbb{E} \left[ Z_D^2 \middle| D \right] \ge \frac{1}{10} \mathbb{E} \left[ Z_D^2 \right] \right) \ge \frac{99\mathbb{E} [Z_D^2]^2}{100 \mathrm{Var} \left( \mathbb{E} \left[ Z_D^2 \middle| D \right] \right)} \ge \frac{99}{100} \frac{4}{5}
\end{align}
Putting this together with \cref{eq:ZD-bound}, conditioning on the event $\left\{ \mathbb{E} \left[ Z_D^2 \middle| D \right] \ge \frac{1}{10} \mathbb{E} [Z_D^2] \right\}$,
\begin{align}
    \frac{99}{100} \frac{4}{5} \mathbb{E} \left[ \mathrm{Pr} \left( Z_D^2 \le \frac{\mathbb{E} [Z_D^2]}{10} \right) \right] \le \mathbb{E} \left[ \mathrm{Pr} \left( Z_D^2 \le \mathbb{E} \left[ Z_D^2 \middle| D \right] \right) \right] \le 1 - \frac{33}{100}
\end{align}
In particular,
\begin{equation}
    \mathbb{E} \left[ \mathrm{Pr} \left( Z_D^2 \le \frac{\mathbb{E} [Z_D^2]}{10} \right) \right] \le \mathbb{E} \left[ \mathrm{Pr} \left( Z_D^2 \le \frac{\mathbb{E} [Z_D^2]}{10} \right) \right] < 0.9
\end{equation}
Finally, we invoke the lower bound on $\mathbb{E} [Z_D^2] \gtrsim \frac{H^{3/2}}{N}$ from \Cref{lemma:LB-E[ZD2]} and use the fact that $Z_D = J(\pi_1^*) - J(\pi_2^*)$ to complete the proof. 
\end{proof}

\paragraph{Proof of \Cref{theorem:lb-known-transition}}
From \Cref{lemma:exp-value-diff}, there exists a constant $C > 0$ such that,
\begin{equation}
    \mathbb{E} \left[ \mathrm{Pr} \left( \left| J_{\mathcal{M}} (\pi^*_1) - J_{\mathcal{M}} (\pi^*_2) \right| \le \frac{C H^{3/2}}{N} \middle| D \right) \right] \le 0.9.
\end{equation}
Therefore, from \Cref{lemma:2sided-1sided} and \Cref{lemma:symmetrization}, we conclude there exists an MDP $\mathcal{M}$ such that,
\begin{equation}
    \mathrm{Pr} \left( J_{\mathcal{M}} (\pi^*) - J_{\mathcal{M}} (\widehat{\pi}) \ge \frac{C H^{3/2}}{N} \right) \ge \frac{1-0.95}{2} = 0.025. 
\end{equation}

\subsection{Proof of Theorem~\ref{thm.4statemdpunbiased}}\label{app.4stateunbiasedproof}

Consider a Poissonized setting where we receive $\textrm{Poi}(N)$ trajectories\footnote{We can always simulate Poisson sampling with $\mathrm{Poi}(N/2)$ trajectories based on $N$ trajectories sampled based on the multinomial distribution, and the failure probability is exponentially small. }. Let $X_t$ represent the number of trajectories in which the expert visits state 2 at time $t$ in the dataset. Under the Poisson setting, $\{X_t:1\leq t\leq H\}$ are mutually independent with each following distribution $\textrm{Poi}(N w_t)$, where $w_t = (1-1/N)^{t-1}/N = \mathrm{Pr}_{\pi^*}(s_t = 2) $ since this probability does not depend on what $\pi^*$ is.  

Let $\Delta = \lfloor c \log (NH) \rfloor$, where $c>0$ is some constant to be determined later. Given random observations $\{X_t:1\leq t\leq H\}$, the policy we output is:
\begin{align}
    \widehat{\pi}_t(\textrm{red}\mid 2) = \begin{cases}  \frac{\sum_{i = (k-1)\Delta+1}^{k\Delta} X_i U_i}{\sum_{i = (k-1)\Delta+1}^{k\Delta}X_i} &
\text{ if } (k-1)\Delta < t \leq k\Delta   \\
1 & \sum_{i = (k-1)\Delta+1}^{k\Delta}X_i =0 
    \end{cases}
\end{align}

At any time $t$, the expert has marginal probability on state $3$, $\mathrm{Pr}_{\pi^*}(s_t = 3) = \sum_{i = 1}^{t-1} w_i U_i$, while our policy $\widehat{\pi}$ has expected marginal probability $\mathbb{E}[\mathrm{Pr}_{\widehat{\pi}}(s_t = 3) ] = \sum_{i = 1}^{t-1} w_i \mathbb{E}[\widehat{\pi}_t(\text{red} \mid 2)]$. We aim to show that
\begin{align}\label{eqn.eachstatemarginal}
    \sum_{t = 1}^H |\sum_{i = 1}^{t-1} w_i \mathbb{E}[\widehat{\pi}_t(\text{red}\mid 2)] -  \sum_{i = 1}^{t-1} w_i U_i|\lesssim \frac{H\log( HN)}{N}.
\end{align}
Using the property that for $X \sim \mathrm{Poi} (\mu)$ and independent $Y \sim \mathrm{Poi} (\lambda)$, $\mathbb{E} \left[ \frac{X}{X+Y} \middle| X+Y > 0 \right] = \frac{\mu}{\mu + \lambda}$, we know that if $(k-1)\Delta <t\leq k \Delta$, \begin{align}\label{eqn.eachtnearlygood}
    \left| \mathbb{E}[\widehat{\pi}_t(\text{red}\mid 2)] -  \frac{\sum_{i = (k-1)\Delta+1}^{k\Delta} w_i U_i}{\sum_{i = (k-1)\Delta+1}^{k\Delta}w_i} \right |\leq \mathrm{Pr} \left(\sum_{i = (k-1)\Delta+1}^{k\Delta}X_i = 0 \right) \lesssim \frac{1}{N^2H^2}
\end{align}
if we take $c$ to be large enough constant. Clearly
\begin{align}
   \left| \left( \sum_{i = 1}^{t-1} w_i \right) \frac{\sum_{i = 1}^\Delta w_iU_i}{\sum_{i = 1}^\Delta w_i} - \sum_{i =1}^{t-1} w_i U_i \right| \lesssim \frac{\log (HN)}{N},
\end{align}
for $1\leq t\leq \Delta$, where we used the fact that each $0\leq w_i \lesssim 1/N$. Indeed, now the total bias is upper bounded by $H\frac{\log (HN)}{N} + \frac{1}{N^2H^2} H^2 \lesssim H\frac{\log (HN)}{N}$ once we combine it with~(\ref{eqn.eachtnearlygood}).

Since the marginal distributions of states 1 and 2 do not depend on the policy, we have just shown that for every time $t$ and every state $s$, 
\begin{align}
    |\mathbb{E}[\mathrm{Pr}_{\widehat{\pi}}(s_t = s)] -\mathrm{Pr}_{\pi^*}(s_t = s)| \lesssim \frac{\log (NH)}{N},
\end{align}
which implies the final result.

%------------------------------------------------------------------------
%------------------------------------------------------------------------
%------------------------------------------------------------------------
%------------------------------------------------------------------------

\subsection{Proof of Corollary~\ref{cor.3stategeneral}}

We use sample splitting and cut $N$ trajectories into two halves. We construct the states $\mathcal{S}_0$ required by the \textsc{Mimic-Mixture} algorithm to be the states that are visited in the first half of the dataset, so we know the expert actions there. Then, we search from the last layer backwards for the first time that there exists one state that was not observed in the first half of the dataset. Denote that time as $t_0$. If there are two states in time $t_0$ that are unseen, then following the arguments in the binary state case in the proof of Theorem~\ref{thm.mimimdefficient}, we know that \textsc{Mimic-MD} already works; if there exists only one state that is unseen at time $t_0$, we use \textsc{Mimic-Mixture} to construct the nearly unbiased estimator of the state-action marginal distribution for one of the other two states. 

The final result can be proved upon noticing the following two observations. First, by the data processing inequality~(Lemma~\ref{lemma:TVshrink}), if we have nearly unbiased estimator at time $t_0$, we have nearly unbiased estimator at time $H$. Second, the unseen state at time $t_0$ must have marginal probability at most $\widetilde{O}(1/N)$ due Binomial concentration, which implies that once we construct the $\pi^{\textrm{L}}$ and $\pi^{\textrm{S}}$ policies in \textsc{Mimic-Mixture} as the two policies that maximize/minimize the marginal probability of the target state while guaranteeing this unseen state has expected visitation probability at most $\widetilde{O}(1/N)$, the overall suboptimality at time $H$ is at most $\widetilde{O}(1/N)$.

\subsection{Proof of Lemma \ref{lemma:unbiased_coefficients}}

We first show that if the coefficients $\beta^*(\mathsf{tr}), \beta^{\textrm{S}}(\mathsf{tr}),\beta^{\textrm{L}}(\mathsf{tr}) \in [0,1]$ such that all the unbiasedness, order, and feasibility properties hold, then we can construct a policy $\widehat{\pi}$ such that whose expected state visitation probability at the terminal state $s^*$ is close to that of the expert up to $O(1/N)$. 

The choice of the mixing coefficient $\widehat{\alpha}$ is slightly different from the approach of plugging \eqref{eq:linear_est} into \eqref{eq:mixing_coefficient_ideal}: for each $\mathsf{tr}$, we subsample the Poisson random variable $X(\mathsf{tr})$ using the subsampling probability $\beta^{\textrm{L}}(\mathsf{tr}) - \beta^{\textrm{S}}(\mathsf{tr})\in [0,1]$ to arrive at another Poisson random variable $Y(\mathsf{tr})$, and further subsample $Y(\mathsf{tr})$ with probability $(\beta^*(\mathsf{tr}) - \beta^{\textrm{S}}(\mathsf{tr}))/(\beta^{\textrm{L}}(\mathsf{tr}) - \beta^{\textrm{S}}(\mathsf{tr}))\in [0,1]$ to obtain a third Poisson random variable $Z(\mathsf{tr})$. The subsamplings for different $\mathsf{tr}$ are mutually independent. Then we construct the mixing coefficient by taking ratios as in \eqref{eq:mixing_coefficient}.

By the subsampling property of Poisson random variables and the mutual independence of $\{X(\mathsf{tr}) \}$, the Poisson random variables $Z(\mathsf{tr})\sim \Poi(N/2\cdot (\beta^*(\mathsf{tr}) - \beta^{\textrm{S}}(\mathsf{tr}))\mathrm{Pr}_{\pi^*}(\mathsf{tr}))$ and $Y(\mathsf{tr}) - Z(\mathsf{tr})\sim \Poi(N/2\cdot (\beta^{\textrm{L}}(\mathsf{tr}) - \beta^*(\mathsf{tr}))\mathrm{Pr}_{\pi^*}(\mathsf{tr}))$ are independent. Since for independent $X\sim \Poi(\lambda), Y\sim \Poi(\mu)$, it holds that
\begin{align*}
\bE\left[\frac{X}{X+Y} ~ \bigg| ~ X+Y\neq 0\right] = \frac{\lambda}{\lambda+\mu}, 
\end{align*}
it is clear that the mixing coefficient $\widehat{\alpha}$ constructed in \eqref{eq:mixing_coefficient} satisfies
\begin{align*}
\bE\left[\widehat{\alpha} ~ \bigg| ~ \sum_{\mathsf{tr}} Y(\mathsf{tr}) \neq 0 \right] &= \bE\left[\frac{\sum_{\mathsf{tr}} Z(\mathsf{tr}) }{\sum_{\mathsf{tr}} Y(\mathsf{tr})} ~ \bigg| ~ \sum_{\mathsf{tr}} Y(\mathsf{tr}) \neq 0 \right] \\
&= \frac{\sum_{\mathsf{tr}} (\beta^*(\mathsf{tr}) - \beta^{\textrm{S}}(\mathsf{tr}))\mathrm{Pr}_{\pi^*}(\mathsf{tr})}{\sum_{\mathsf{tr}} (\beta^{\textrm{L}}(\mathsf{tr}) - \beta^{\textrm{S}}(\mathsf{tr}))\mathrm{Pr}_{\pi^*}(\mathsf{tr})} \\
&\overset{(i)}{=} \frac{\mathrm{Pr}_{\pi^*}(s_t = s^*) - \mathrm{Pr}_{\pi^{\textrm{S}}}(s_t = s^*)}{ \mathrm{Pr}_{\pi^{\textrm{L}}}(s_t = s^*) -  \mathrm{Pr}_{\pi^{\textrm{S}}}(s_t = s^*)} = \alpha^*, 
\end{align*}
where (i) is due to the unbiasedness requirement for the coefficients. Consequently, 
\begin{align}\label{eq:bias_upper_bound}
| \bE[\widehat{\alpha}] - \alpha^* | \le \mathrm{Pr}\left(\sum_{\mathsf{tr}} Y(\mathsf{tr}) = 0 \right) &= \exp\left( -\frac{N}{2} \sum_{\mathsf{tr}} (\beta^{\textrm{L}}(\mathsf{tr}) - \beta^{\textrm{S}}(\mathsf{tr}))\mathrm{Pr}_{\pi^*}(\mathsf{tr})\right) \nonumber\\
&= \exp\left(-\frac{N}{2}(\mathrm{Pr}_{\pi^{\textrm{L}}}(s_t = s^*) - \mathrm{Pr}_{\pi^{\textrm{S}}}(s_t = s^*))\right). 
\end{align}
Using \eqref{eq:bias_upper_bound} and the definition of $\widehat{\pi} = \widehat{\alpha}\pi^{\textrm{L}} + (1-\widehat{\alpha})\pi^{\textrm{S}}$, the bias in Theorem \ref{theorem:single_state} satisfies
\begin{align}\label{eq:bias_upper_bound_final}
&|\bE[\mathrm{Pr}_{\widehat{\pi}}(s_t = s^*)] - \mathrm{Pr}_{\pi^*}(s_t = s^*)| \nonumber\\
&= |\bE[\widehat{\alpha}](\mathrm{Pr}_{\pi^{\textrm{L}}}(s_t = s^*) - \mathrm{Pr}_{\pi^{\textrm{S}}}(s_t = s^*)) - (\mathrm{Pr}_{\pi^*}(s_t = s^*) - \mathrm{Pr}_{\pi^{\textrm{S}}}(s_t = s^*)) | \nonumber\\
&= |\bE[\widehat{\alpha}] - \alpha^*|\cdot (\mathrm{Pr}_{\pi^{\textrm{L}}}(s_t = s^*) - \mathrm{Pr}_{\pi^{\textrm{S}}}(s_t = s^*)) \nonumber\\
&\le  \exp\left(-\frac{N}{2}(\mathrm{Pr}_{\pi^{\textrm{L}}}(s_t = s^*) - \mathrm{Pr}_{\pi^{\textrm{S}}}(s_t = s^*))\right)\cdot (\mathrm{Pr}_{\pi^{\textrm{L}}}(s_t = s^*) - \mathrm{Pr}_{\pi^{\textrm{S}}}(s_t = s^*)) \nonumber\\
&\le \frac{2}{eN}, 
\end{align}
where the last inequality is due to $\sup_{x} xe^{-t x} = 1/(et)$ for any $t>0$. According to \eqref{eq:bias_upper_bound_final}, the claimed bias upper bound in Theorem \ref{theorem:single_state} is proved. 

We introduce several useful notations for the proof. Although the state space $\calS$ is shared among all times $t\in [H]$, we use the notation $\calS_t$ to denote the state space at time $t$. Given a policy $\pi$, for $t_1<t_2$ and states $s_{t_1}\in \calS_{t_1}, s_{t_2}\in \calS_{t_2}$, we use $\mathrm{Pr}_\pi(s_{t_1}\to s_{t_2})$ to denote the probability of reaching $s_{t_2}$ from $s_{t_1}$ through \emph{only the states in $\calS_0$} under the policy $\pi$; in other words,
\begin{align}\label{eq:prob-t1-to-t2}
\mathrm{Pr}_\pi(s_{t_1}\to s_{t_2}) \triangleq \bE\left[\sum_{s_{t_1+1}\in \calS_{t_1+1}\cap \calS_0 }\cdots \sum_{s_{t_2-1}\in \calS_{t_2-1} \cap \calS_0 } \prod_{t=t_1}^{t_2-1} P_{t}(s_{t+1}\mid s_{t}, \pi(s_{t}))\right], 
\end{align}
where the expectation is taken with respect to the possible randomness in the policy $\pi$. When we start from the initial state distribution $\rho$, we also write $\mathrm{Pr}_\pi(\mathsf{s}\to s_t)$ to denote
\begin{align}\label{eq:prob-s-to-t}
\mathrm{Pr}_\pi(\mathsf{s}\to s_t) = \sum_{s_1\in\calS_1\cap \calS_0} \rho(s_1)\cdot \mathrm{Pr}_\pi(s_1\to s_t), 
\end{align}
where $\mathsf{s}$ denotes ``start''. Similarly, we also define the probability from $s_t$ to the end by
\begin{align}\label{eq:prob-t-to-f}
\mathrm{Pr}_\pi(s_t \to \mathsf{f}) = \sum_{s_H \in \calS_H} \mathrm{Pr}_\pi(s_t \to s_H),
\end{align}
where $\mathsf{f}$ denotes ``finish''. Note the following difference between \eqref{eq:prob-s-to-t} and our usual notation $\mathrm{Pr}_\pi(s_t =s)$: the latter quantity does not require that the trajectory to $s_t$ only consists of states in $\calS_0$. The main motivation behind \eqref{eq:prob-t1-to-t2}, \eqref{eq:prob-s-to-t}, and \eqref{eq:prob-t-to-f} is that $\mathrm{Pr}_{\pi^*}(\rho\to s_t)$ is known to the learner solely based on the publicly known expert actions at states $\calS_0$, and the probabilities $\mathrm{Pr}_{\pi^*}(s_{t_1}\to s_{t_2})$ and $\mathrm{Pr}_{\pi^*}(s_{t_1}\to\mathsf{f})$ are known as long as the learner knows the first action $\pi^*(s_{t_1})$. 

We also combine and partition all trajectories $\mathsf{tr}\in \calS^H$ into several disjoint groups. For a given trajectory $\mathsf{tr} = (s_1,\cdots,s_H)$, we define the following notations: 
\begin{enumerate}
	\item The \emph{characteristic} of $\mathsf{tr}$, or $\mathsf{c}(\mathsf{tr})$, is the set of all times (except for $t=H$) and the corresponding states in the trajectory which are not in $\calS_0$. Mathematically, $\mathsf{c}(\mathsf{tr})=\{(t, s_t): t\in [H-1], s_t\notin \calS_0 \}$. 
	\item The \emph{starting point} of $\mathsf{c}$, or $t_\ell(\mathsf{c})$, is defined to be the smallest $t\in [H]$ with $(t,s_t)\in \mathsf{c}$ for some $s_t\in \calS_t$. If $\mathsf{c} = \emptyset$, we define $t_\ell(\mathsf{c}) = \bot$. 
	\item The \emph{ending point} of $\mathsf{c}$, or $t_r(\mathsf{c})$, is defined to be the largest $t\in [H]$ with $(t,s_t)\in \mathsf{c}$ for some $s_t\in \calS_t$. If $\mathsf{c} = \emptyset$, we define $t_r(\mathsf{c}) = \bot$. 
	\item For each possible characteristic $\mathsf{c}$, let the \emph{$\mathsf{c}$-group}, or $\calT_{\mathsf{c}}$, be the set of all trajectories $\mathsf{tr}$ with $\mathsf{c}(\mathsf{tr})=\mathsf{c}$. 
	\item For each possible pair $(t,s_t)$ with $t\in [H-1], s_t\in \calS_t \backslash \calS_0$, let $\calG_{t,s_t}$ be the set of all characteristics $\mathsf{c}$ such that $t_{\ell}(\mathsf{c})=t$ and $(t,s_t)\in \mathsf{c}$. The set $\calG_\bot$ is defined analogously. 
\end{enumerate}
The main idea behind the above notations is that we require the dependence of coefficients $\beta^\dagger(\mathsf{tr})$ on $\mathsf{tr}$ only through $\mathsf{c}(\mathsf{tr})$, for all $\dagger\in\{*, \textrm{L}, \textrm{S} \}$, and therefore we only need to specify the coefficients for every $\mathsf{c}$-group. Consequently, we denote by $\beta^\dagger(\calT_{\mathsf{c}})$ the common coefficient $\beta^\dagger(\mathsf{tr})$ for all $\mathsf{tr}\in \calT_{\mathsf{c}}$, and also by $X(\calT_{\mathsf{c}}) = \sum_{\mathsf{tr}\in \calT_{\mathsf{c}}} X(\mathsf{tr})$ the total Poisson count for $\calT_{\mathsf{c}}$. It is clear that for $\mathsf{c} = \{(t_i,s_{t_i}): i\in [m] \}$, we have
\begin{align}\label{eq:prob-c-group}
\mathrm{Pr}_\pi(\calT_{\mathsf{c}}) = \mathrm{Pr}_\pi (\mathsf{s}\to s_{t_1})\cdot \prod_{j=1}^{m-1} \mathrm{Pr}_\pi(s_{t_j} \to s_{t_{j+1}}) \cdot \mathrm{Pr}_\pi (s_{t_m}\to \mathsf{f}),
\end{align}
and $X(\calT_{\mathsf{c}}) \sim \Poi(N/2 \cdot \mathrm{Pr}_{\pi^*}(\calT_{\mathsf{c}}))$. Note that the first probability term $\mathrm{Pr}_\pi (\mathsf{s}\to s_{t_1})$ of \eqref{eq:prob-c-group} in fact does not depend on $\pi$, in the sequel the following notation will also be useful: 
\begin{align}\label{eq:prob-tilde-c-group}
\widetilde{\mathrm{Pr}}_\pi(\calT_{\mathsf{c}}) =  \prod_{j=1}^{m-1} \mathrm{Pr}_\pi(s_{t_j} \to s_{t_{j+1}}) \cdot \mathrm{Pr}_\pi (s_{t_m}\to \mathsf{f}). 
\end{align}

The starting point $t_\ell(\mathsf{c})$, as well as the group $\calG_{t,s_t}$, is used for further partitioning the $\mathsf{c}$-groups. Specifically, we sequentially assign the coefficients $\beta^\dagger(\calT_{\mathsf{c}})$ to all characteristics in each group $\calG$ via an appropriate order, and aim to show that the following three conditions hold for each $\calG\in \{\calG_{t,s_t}: t\in [H], s_t\in \calS_t \backslash\calS_0 \}\cup \{\calG_\bot\}$ (without loss of generality we assume that the target state $s^*$ belongs to the last layer, i.e. $s^*\in \calS_H$): 
\begin{enumerate}
	\item Unbiasedness: for $\calG \neq \calG_\bot$, it holds that
	\begin{align}\label{eq:cond_unbiased}
	  \sum_{\mathsf{c} \in \calG} \beta^\dagger(\calT_{\mathsf{c}})\cdot \widetilde{\mathrm{Pr}}_{\pi^*}(\calT_{\mathsf{c}}) = \sum_{\mathsf{c}\in \calG} \widetilde{\mathrm{Pr}}_{\pi^\dagger}(\calT_{\mathsf{c}})\cdot \frac{\mathrm{Pr}_{\pi^\dagger}(s_{t_r(\mathsf{c})}\to s^*) }{ \mathrm{Pr}_{\pi^\dagger}(s_{t_r(\mathsf{c})}\to \mathsf{f})}, \quad \dagger\in \{*, \textrm{L}, \textrm{S} \}, 
	\end{align}
	where we recall that $t_r(\mathsf{c})$ is the ending point of $\mathsf{c}$. For $\calG = \calG_\bot$ and $\mathsf{c} = \emptyset$, the condition \eqref{eq:cond_unbiased} is replaced by
	\begin{align}\label{eq:cond_unbiased_special}
	 \beta^\dagger(\calT_{\emptyset}) \cdot \mathrm{Pr}_{\pi^*}(\calT_{\emptyset}) = \mathrm{Pr}_{\pi^\dagger}(\mathsf{s}\to s^*). 
	\end{align}
	\item Order: it always holds that $0\le \beta^{\textrm{S}}(\calT_{\mathsf{c}}) \le \beta^*(\calT_{\mathsf{c}})\le \beta^{\textrm{L}}(\calT_{\mathsf{c}})\le 1$ for all $\mathsf{c}\in \calG$.
	\item Feasibility: for $\dagger\in \{*,\textrm{L}, \textrm{S} \}$, the coefficient $\beta^\dagger(\calT_{\mathsf{c}})$ only depends on the public information and $\{\pi^*(s_{t_j})\}_{j\in [m]}$, where $\mathsf{c} = \{(t_j, s_{t_j}): j\in [m] \}$. 
\end{enumerate}
Note that the order and feasibility conditions are the same as original ones, and below we show that the above unbiasedness condition implies the original unbiasedness property. For a given $\calG \neq \calG_\bot$, we must have $\calG = \calG_{t,s_t}$ for some $t\in [H], s_t\in \calS_t \backslash \calS_0$. Now multiplying $\mathrm{Pr}_{\pi^*}(\mathsf{s}\to s_{t}) = \mathrm{Pr}_{\pi^\dagger}(\mathsf{s}\to s_{t})$ to both sides of \eqref{eq:cond_unbiased}, and also using \eqref{eq:prob-c-group}, \eqref{eq:prob-tilde-c-group}, we arrive at
	\begin{align}\label{eq:cond_unbiased_weaker}
\sum_{\mathsf{c} \in \calG} \beta^\dagger(\calT_{\mathsf{c}})\cdot {\mathrm{Pr}}_{\pi^*}(\calT_{\mathsf{c}}) = \sum_{\mathsf{c}\in \calG} {\mathrm{Pr}}_{\pi^\dagger}(\calT_{\mathsf{c}})\cdot \frac{\mathrm{Pr}_{\pi^\dagger}(s_{t_r(\mathsf{c})}\to s^*) }{ \mathrm{Pr}_{\pi^\dagger}(s_{t_r(\mathsf{c})}\to \mathsf{f})}, \quad \dagger\in \{*, \textrm{L}, \textrm{S} \}. 
\end{align}
Using \eqref{eq:cond_unbiased_special} and \eqref{eq:cond_unbiased_weaker}, we have
\begin{align*}
\sum_{\mathsf{tr}} \beta^\dagger(\mathsf{tr}) \cdot \mathrm{Pr}_{\pi^*}(\mathsf{tr}) &= \beta^\dagger(\calT_{\emptyset})\cdot \mathrm{Pr}_{\pi^*}(\calT_{\emptyset}) +  \sum_{\calG\neq \calG_\bot} \sum_{\mathsf{c}\in \calG} \beta^\dagger(\calT_{\mathsf{c}})\cdot \mathrm{Pr}_{\pi^*}(\calT_{\mathsf{c}}) \\
&= \mathrm{Pr}_{\pi^\dagger}(\mathsf{s}\to s^*) + \sum_{\mathsf{c}\neq \emptyset} \mathrm{Pr}_{\pi^\dagger}(\calT_{\mathsf{c}})\cdot \frac{\mathrm{Pr}_{\pi^\dagger}(s_{t_r(\mathsf{c})}\to s^*) }{ \mathrm{Pr}_{\pi^\dagger}(s_{t_r(\mathsf{c})}\to \mathsf{f})} \\
&= \mathrm{Pr}_{\pi^\dagger}(\mathsf{s}\to s^*) + \sum_{ \mathsf{c} = \{(t_j, s_{t_j}): j\in [m] \} } \mathrm{Pr}_{\pi^\dagger} (\mathsf{s}\to s_{t_1})\cdot \prod_{j=1}^{m-1} \mathrm{Pr}_{\pi^\dagger}(s_{t_j} \to s_{t_{j+1}}) \cdot \mathrm{Pr}_{\pi^\dagger} (s_{t_m}\to s^*)\\
&= \mathrm{Pr}_{\pi^\dagger}(s_H = s^*),
\end{align*}
where the last identity follows from the partition of all trajectories to $s^*$ into disjoint characteristics. This is exactly the original unbiasedness property. 

Next we show that for each $\calG$ we could fulfill the above conditions. We will first deal with the group $\calG_\bot$ in a special way, and then handle other groups $\calG_{t,s_t}$ by induction on $t=H-1,H-2,\cdots,1$. 
\begin{remark}
Note that the condition \eqref{eq:cond_unbiased} cannot be replaced by \eqref{eq:cond_unbiased_weaker} in general, as it might happen that $\mathrm{Pr}_{\pi^\dagger}(\calT_{\mathsf{c}}) = 0$ while $\widetilde{\mathrm{Pr}}_{\pi^\dagger}(\calT_{\mathsf{c}}) > 0$. For example, in the special case of $\calS_0 = \emptyset$, the condition \eqref{eq:cond_unbiased_weaker} is totally non-informative for $\calG = \calG_{t,s_t}$ with $t\ge 2$. 

We also remark that the coefficient $\beta^\dagger(\calT_{\mathsf{c}})$ must be constructed for every characteristic $\mathsf{c}$, even if for certain $\calS_0$ there does not exist a trajectory $\mathsf{tr}$ such that $\mathsf{c} = \mathsf{c}(\mathsf{tr})$ (e.g. consider $\calS_0 = \emptyset$). This is because our construction is sequentially inductive, and therefore must be done step after step. 
\end{remark} 

\vspace{0.2cm}
\noindent \textbf{Edge case: $\calG = \calG_\bot$.} In this case, the only element of $\calG_\bot$ is $\mathsf{c} = \emptyset$, and we have
\begin{align*}
&\mathrm{Pr}_{\pi^*}(\calT_{\emptyset}) = \mathrm{Pr}_{\pi^*}(\mathsf{s}\to \mathsf{f}) = \mathrm{Pr}_{\pi^{\textrm{L}}}(\mathsf{s}\to \mathsf{f}) = \mathrm{Pr}_{\pi^{\textrm{S}}}(\mathsf{s}\to \mathsf{f}) \\
&\mathrm{Pr}_{\pi^*}(\mathsf{s}\to s^*) = \mathrm{Pr}_{\pi^{\textrm{L}}}(\mathsf{s}\to s^*) = \mathrm{Pr}_{\pi^{\textrm{S}}}(\mathsf{s}\to s^*), 
\end{align*}
and all above quantities are publicly known. By \eqref{eq:cond_unbiased_special}, all three conditions are fulfilled by choosing
\begin{align*}
\beta^*(\calT_{\emptyset}) = \beta^{\textrm{L}}(\calT_{\emptyset}) = \beta^{\textrm{S}}(\calT_{\emptyset}) = \frac{\mathrm{Pr}_{\pi^*}(\mathsf{s}\to s^*)}{\mathrm{Pr}_{\pi^*}(\mathsf{s}\to \mathsf{f})}\in [0,1]. 
\end{align*}

\vspace{0.2cm}
\noindent \textbf{Base step of induction: $\calG = \calG_{H-1,s_{H-1}}$ for some $s_{H-1}\in\calS_{H-1}$.} In this case, the set $\calG_{H-1,s_{H-1}}$ has a unique element $\mathsf{c} = \{(H-1,s_{H-1}) \}$, and the condition \eqref{eq:cond_unbiased} is equivalent to
\begin{align}\label{eq:coeff_base}
\beta^\dagger(\calT_{\mathsf{c}}) = \mathrm{Pr}_{\pi^\dagger}(s_{H-1}\to s^*), \quad \dagger\in \{*, \textrm{L}, \textrm{S} \},
\end{align}
as $\widetilde{\mathrm{Pr}}_{\pi^*}(\calT_{\mathsf{c}}) = \widetilde{\mathrm{Pr}}_{\pi^\dagger}(\calT_{\mathsf{c}}) = \mathrm{Pr}_{\pi^\dagger}(s_{H-1}\to \mathsf{f}) = 1$. By definition of the extremal policies $\pi^{\textrm{L}}$ and $\pi^{\textrm{S}}$, Lemma \ref{lemma:extremal_policy} at the end of the section shows that the choice in \eqref{eq:coeff_base} also satisfies the order condition. Finally, the probability in \eqref{eq:coeff_base} for $\dagger\in \{\textrm{L}, \textrm{S}\}$ is determined by the known transition and extremal policies, while for $\dagger = *$, the coefficient only requires the additional information $\pi^*(s_{H-1})$. As the characteristic $\mathsf{c}$ is $\{(H-1,s_{H-1})\}$, the choice of \eqref{eq:coeff_base} also satisfies the feasibility condition.  

\vspace{0.2cm}
\noindent \textbf{Inductive step: $\calG = \calG_{t,s_t}$ after handling all $\calG_{t',s_{t'}}$ for $t'>t$.} We choose the coefficients $\beta^\dagger(\calT_{\mathsf{c}})$ with $\calT_{\mathsf{c}}\in \calG_{t,s_t}$ for each $\dagger\in \{*, \textrm{L}, \textrm{S} \}$, respectively. 

The choice for $\dagger = *$ is the simplest, and is given by
\begin{align}\label{eq:coeff_ind_*}
\beta^*(\calT_{\mathsf{c}}) \triangleq \frac{\mathrm{Pr}_{\pi^*}(s_{t_r(\mathsf{c})}\to s^*) }{ \mathrm{Pr}_{\pi^*}(s_{t_r(\mathsf{c})}\to \mathsf{f})}\in [0,1], \quad \forall \mathsf{c} \in \calG_{t,s_t}. 
\end{align}
Plugging \eqref{eq:coeff_ind_*} into \eqref{eq:cond_unbiased}, it is clear that the unbiased condition holds for $\dagger = *$. Moreover, both the numerator and the denominator in \eqref{eq:coeff_ind_*} only require the additional information $\pi^*(s_{t_r(\mathsf{c})})$, and thus $\beta^*$ satisfies the feasibility condition. 

Next we construct $\beta^{\textrm{L}}(\calT_{\mathsf{c}})$ such that the unbiased and feasibility conditions hold, with $\beta^{\textrm{L}}(\calT_{\mathsf{c}})\in [\beta^*(\calT_{\mathsf{c}}) ,1]$. An entirely symmetric argument also leads to the claimed construction of $\beta^{\textrm{S}}(\calT_{\mathsf{c}})$. For every $\mathsf{c}\in \calG_{t,s_t}$, the coefficient $\beta^{\textrm{L}}(\calT_{\mathsf{c}})$ is chosen to be
\begin{align}\label{eq:coeff_ind_L}
\beta^{\textrm{L}}(\calT_{\mathsf{c}}) = (1-\alpha)\beta_0^{\textrm{L}}(\calT_{\mathsf{c}}) + \alpha \beta_1^{\textrm{L}}(\calT_{\mathsf{c}}),
\end{align}
with some scalar $\alpha\in [0,1]$ independent of $\mathsf{c}$, and the candidate coefficients $\beta_0^{\textrm{L}}, \beta_1^{\textrm{L}}$ are defined as
\begin{align*}
\beta_0^{\textrm{L}}(\calT_{\mathsf{c}}) \equiv 1, \qquad \beta_1^{\textrm{L}}(\calT_{\mathsf{c}}) = \begin{cases}
\beta^*(\calT_{\mathsf{c}}) & \text{if } \mathsf{c} = \{(t,s_t)\}, \\
\beta^{\textrm{L}}(\calT_{\mathsf{c} \backslash \{(t,s_t) \} }) & \text{otherwise.}
\end{cases}
\end{align*}
We first show that both $\beta_0^{\textrm{L}}, \beta_1^{\textrm{L}}$ satisfy the feasibility condition. This result is trivial for the constant $\beta_0^{\textrm{L}}$; the coefficient $\beta_1^{\textrm{L}}$ is also feasible, for both coefficients $\beta^*(\calT_{\mathsf{c}})$ in \eqref{eq:coeff_ind_*} and $\beta^{\textrm{L}}(\calT_{\mathsf{c} \backslash \{(t,s_t) \} })$ in the inductive hypothesis are feasible. Moreover, whenever $\mathsf{c}\in \calG_{t,s_t}$ is not a singleton, by the inductive hypothesis and \eqref{eq:coeff_ind_*} we have
\begin{align*}
 \beta^*(\calT_{\mathsf{c}}) = \frac{\mathrm{Pr}_{\pi^*}(s_{t_r(\mathsf{c})}\to s^*) }{ \mathrm{Pr}_{\pi^*}(s_{t_r(\mathsf{c})}\to \mathsf{f})} =  \frac{\mathrm{Pr}_{\pi^*}(s_{t_r(\mathsf{c} \backslash \{(t,s_t)\} )}\to s^*) }{ \mathrm{Pr}_{\pi^*}(s_{t_r(\mathsf{c}\backslash \{(t,s_t)\})}\to \mathsf{f})}  =  \beta^*(\calT_{\mathsf{c} \backslash \{(t,s_t) \} }) \le \beta^{\textrm{L}}(\calT_{\mathsf{c} \backslash \{(t,s_t) \} }) \le 1. 
\end{align*}
Consequently, we always have $\beta_0^{\textrm{L}}, \beta_1^{\textrm{L}}\in [\beta^*, 1]$, therefore the order condition $\beta^{\textrm{L}} \in [\beta^*, 1]$ holds for any mixture in \eqref{eq:coeff_ind_L}. 

Now it remains to show that there exists $\alpha\in [0,1]$ such that the mixture $\beta^{\textrm{L}}$ in \eqref{eq:coeff_ind_L} satisfies the condition \eqref{eq:cond_unbiased} for $\dagger = \textrm{L}$, and that this common value $\alpha$ is feasible with respect to all $\mathsf{c} \in \calG_{t,s_t}$. For the first claim, it suffices to prove that
\begin{align}
A\triangleq \sum_{\mathsf{c}\in \calG_{t,s_t}} \beta_0^{\textrm{L}}(\calT_{\mathsf{c}}) \cdot \widetilde{\mathrm{Pr}}_{\pi^*}(\calT_{\mathsf{c}}) &\ge \sum_{\mathsf{c}\in \calG_{t,s_t}} \widetilde{\mathrm{Pr}}_{\pi^{\textrm{L}}}(\calT_{\mathsf{c}})\cdot \frac{\mathrm{Pr}_{\pi^{\textrm{L}}}(s_{t_r(\mathsf{c})}\to s^*) }{ \mathrm{Pr}_{\pi^{\textrm{L}}}(s_{t_r(\mathsf{c})}\to \mathsf{f})} \triangleq C, \label{eq:candidate_1} \\
B\triangleq \sum_{\mathsf{c}\in \calG_{t,s_t}} \beta_1^{\textrm{L}}(\calT_{\mathsf{c}}) \cdot \widetilde{\mathrm{Pr}}_{\pi^*}(\calT_{\mathsf{c}}) &\le \sum_{\mathsf{c}\in \calG_{t,s_t}} \widetilde{\mathrm{Pr}}_{\pi^{\textrm{L}}}(\calT_{\mathsf{c}})\cdot \frac{\mathrm{Pr}_{\pi^{\textrm{L}}}(s_{t_r(\mathsf{c})}\to s^*) }{ \mathrm{Pr}_{\pi^{\textrm{L}}}(s_{t_r(\mathsf{c})}\to \mathsf{f})} = C. \label{eq:candidate_2}
\end{align}
Given \eqref{eq:candidate_1} and \eqref{eq:candidate_2}, the parameter $\alpha\in [0,1]$ to fulfill the unbiased condition \eqref{eq:cond_unbiased} is
\begin{align}\label{eq:candidate_mixing}
\alpha = \frac{A-C}{A-B}. 
\end{align}

To establish \eqref{eq:candidate_1}, \eqref{eq:candidate_2} and show that the parameter $\alpha$ in \eqref{eq:candidate_mixing} is feasible, we will find simplified expressions for $A, B$, and $C$. First we claim that
\begin{align}
A &= 1,  \label{eq:A_value} \\ 
C &= \mathrm{Pr}_{\pi^{\text{L}}}(s_H = s^* \mid s_t ) \label{eq:C_value}. 
\end{align}
To show \eqref{eq:A_value}, consider all possible trajectories starting from $s_t$ at time $t$. Partition the trajectories into disjoint sets labeled by different characteristics $\mathsf{c}$, i.e. when and on which states the trajectory hits $\calS_0$. It is clear by \eqref{eq:prob-tilde-c-group} that the probability of the set labeled by $\mathsf{c}$, conditioned on starting from $s_t$ at time $t$, is precisely $\widetilde{\mathrm{Pr}}_{\pi^*}(\calT_{\mathsf{c}})$ under the expert policy $\pi^*$. Summing them up gives $A=1$. The identity \eqref{eq:C_value} could be established in a similar way. 

The quantity $B$ is more complicated to deal with, where a key observation is that $\calG_{t,s_t} \backslash \{(t, s_t)\} = \calG_{\bot}\cup (\cup_{t'>t, s_{t'}\in \calS_{t'} \backslash \calS_0} \calG_{t', s_{t'}}) $. Using the definition of $\beta_1^{\textrm{L}}(\calT_{\mathsf{c}})$, we have
\begin{align*}
B &= \beta^*(\calT_{ \{(t,s_t)\} }) \cdot \mathrm{Pr}_{\pi^*}(s_t \to \mathsf{f}) + \sum_{t'>t}\sum_{s_{t'}\in \calS_{t'}\backslash \calS_0} \sum_{\mathsf{c}' \in \calG_{t',s_{t'}}} \beta^{\textrm{L}}( \calT_{\mathsf{c}'} )\cdot \mathrm{Pr}_{\pi^*}(s_t\to s_{t'})\cdot \widetilde{\mathrm{Pr}}_{\pi^*}(\calT_{\mathsf{c}'}) \\
&\overset{(i)}{=}  \mathrm{Pr}_{\pi^*}(s_t \to s^*) + \sum_{t'>t}\sum_{s_{t'}\in \calS_{t'}\backslash \calS_0}  \mathrm{Pr}_{\pi^*}(s_t\to s_{t'}) \sum_{\mathsf{c}' \in \calG_{t',s_{t'}}} \beta^{\textrm{L}}( \calT_{\mathsf{c}'} )\cdot \widetilde{\mathrm{Pr}}_{\pi^*}(\calT_{\mathsf{c}'}) \\
&\overset{(ii)}{=}  \mathrm{Pr}_{\pi^*}(s_t \to s^*) + \sum_{t'>t}\sum_{s_{t'}\in \calS_{t'}\backslash \calS_0}  \mathrm{Pr}_{\pi^*}(s_t\to s_{t'}) \sum_{\mathsf{c}' \in \calG_{t',s_{t'}}} \widetilde{\mathrm{Pr}}_{\pi^{\textrm{L}}}(\calT_{\mathsf{c}'}) \cdot \frac{\mathrm{Pr}_{\pi^{\textrm{L}}}(s_{t_r(\mathsf{c}')}\to s^*) }{ \mathrm{Pr}_{\pi^{\textrm{L}}}(s_{t_r(\mathsf{c}')}\to \mathsf{f})} \\
&\overset{(iii)}{=} \mathrm{Pr}_{\pi^*}(s_t \to s^*) + \sum_{t'>t}\sum_{s_{t'}\in \calS_{t'}\backslash \calS_0}  \mathrm{Pr}_{\pi^*}(s_t\to s_{t'}) \cdot \mathrm{Pr}_{\pi^{\text{L}}}(s_H = s^* \mid s_{t'} ), 
\end{align*}
where (i) follows from the definition of $\beta^*$ in \eqref{eq:coeff_ind_*}, (ii) uses the inductive hypothesis \eqref{eq:cond_unbiased} for $\calG_{t',s_{t'}}$, and (iii) follows from \eqref{eq:C_value}. In other words, we have
\begin{align}\label{eq:B_value}
B = \mathrm{Pr}_{\pi^* \to \pi^{\textrm{L}} }(s_H = s^* \mid s_t),
\end{align}
where the new policy $\pi^* \to \pi^{\textrm{L}}$ means that starting from $s_t$, the learner initially adopts the policy $\pi^*$ and switches to $\pi^{\text{L}}$ once he visits a state not in $\calS_0$. The expression \eqref{eq:B_value} is obtained by distinguishing the first state not in $\calS_0$ visited by the learner starting from $s_t$. 

By \eqref{eq:A_value}, \eqref{eq:C_value} and \eqref{eq:B_value}, it is clear from Lemma \ref{lemma:extremal_policy} that $A\ge C\ge B$. Regarding the feasibility, it is clear that $A$ and $C$ are both publicly known, and $B$ only requires the knowledge of $\pi^*(s_t)$, which is shared among all $\mathsf{c}\in \calG_{t,s_t}$. Therefore we have completed the inductive step and and the proof of Lemma \ref{lemma:unbiased_coefficients}. 

\begin{lemma}\label[lemma]{lemma:extremal_policy}
For every $t\in [H]$ and $s\in \calS_t$, (proper versions of) the extremal policies $\pi^{\text{\rm L}}, \pi^{\text{\rm S}}$ satisfy
\begin{align*}
\pi^{\text{\rm L}} &\in \underset{\pi \in \Pi_{\mathrm{mimic}}(\calS_0)}{\text{\rm argmax}} \text{\rm Pr}_{\pi}(s_H = s^* \mid s_t = s), \\
\pi^{\text{\rm S}} &\in \underset{\pi \in \Pi_{\mathrm{mimic}}(\calS_0)}{\text{\rm argmin}} \text{\rm Pr}_{\pi}(s_H = s^* \mid s_t = s). 
\end{align*}
\end{lemma}
\begin{proof}
By symmetry we only prove the first claim, and we induct on $t=H-1,H-2,\cdots,1$. For the base case $t=H-1$, the definition of $\pi^{\textrm{L}}$ implies that changing the action $\pi^{\textrm{L}}(s)$ to any $\pi(s)$ cannot decrease $\mathrm{Pr}(s_H = s^*)$, and therefore the statement holds provided that $\mathrm{Pr}_{\pi^{\textrm{L}}}(s_{H-1} = s) > 0$; moreover, in the edge case $\mathrm{Pr}_{\pi^{\textrm{L}}}(s_{H-1} = s) = 0$ we may choose $\pi^{\textrm{L}}(s)$ arbitrarily, so a proper version of $\pi^{\textrm{L}}$ would work. For the induction step, the same local adjustment argument yields to
\begin{align*}
 \text{\rm Pr}_{\pi^{\textrm{L}}}(s_H = s^* \mid s_t = s) &\ge \sum_{s'\in\calS_{t+1}}  \text{\rm Pr}_{\pi}(s_{t+1} = s' \mid s_t = s)  \cdot \text{\rm Pr}_{\pi^{\textrm{L}}}(s_H = s^* \mid s_{t+1} = s') \\
 &\ge \sum_{s'\in\calS_{t+1}}  \text{\rm Pr}_{\pi}(s_{t+1} = s' \mid s_t = s)  \cdot \text{\rm Pr}_{\pi}(s_H = s^* \mid s_{t+1} = s') \\
 &= \text{\rm Pr}_{\pi}(s_H = s^* \mid s_t = s)
\end{align*}
provided that $\mathrm{Pr}_{\pi^{\textrm{L}}}(s_t = s) > 0$, where the second inequality makes use of the induction hypothesis. The edge case is again handled by considering a proper version of $\pi^{\textrm{L}}$. 
\end{proof}

%-----------------------------------------------------------------------
%-----------------------------------------------------------------------
%-----------------------------------------------------------------------